\documentclass{article} 
\usepackage{arxiv}
\usepackage{natbib}
\usepackage{hyperref}
\usepackage{url}
\usepackage{algorithm}
\usepackage{algorithmic}
\usepackage{xcolor}
\usepackage{xspace}
\usepackage{multirow}
\usepackage{tcolorbox}

\usepackage[utf8]{inputenc} %
\usepackage[T1]{fontenc}    %
\usepackage{url}            %
\usepackage{booktabs}       %
\usepackage{amsfonts}       %
\usepackage{nicefrac}       %
\usepackage{microtype}      %
\usepackage{amssymb} 
\usepackage{amsmath,csquotes,xcolor, array,amsthm,thmtools,amsbsy}
\usepackage{xparse}
\usepackage{footmisc}
\usepackage{thm-restate}
\usepackage{mathtools}
\usepackage{tabularx}

\usepackage[shortlabels]{enumitem}

\usepackage[ruled,vlined, algo2e]{algorithm2e} 
\usepackage{algorithm}
\usepackage{color}
\usepackage{graphicx}
\usepackage{subcaption}
\graphicspath{{plots/}}
\newlength{\smallfigwidth}
\newlength{\smallfigheight}
\newlength{\smallfigsep}
\newlength{\legendheight}
\newcommand{\fanar}{\texttt{Fanar1-9b}\xspace}
\newcommand{\qwen}{\texttt{Qwen2.5-7B}\xspace}
\newcommand{\gemma}{\texttt{Gemma3-12B}\xspace}
\newcommand{\oss}{\texttt{gpt-oss-20B}\xspace}

\newcommand{\hotpot}{HotpotQA\xspace}
\newcommand{\NQ}{Natural Questions\xspace}

\newcommand{\mathqa}{Math\xspace}
\newcommand{\truthfulqa}{TruthfulQA\xspace}
\newcommand{\triviaqa}{TriviaQA\xspace}

\def\v+#1{\ensuremath{\mathbf{#1}}\xspace}
\def\m+#1{\ensuremath{\mathbf{#1}}\xspace}

\newcommand{\spara}[1]{\smallskip\noindent{\textbf{#1}}}

\newcommand{\woc}{\ensuremath{\texttt{WOC}}\xspace}
\newcommand{\wcc}{\ensuremath{\texttt{WCC}}\xspace} 
\newcommand{\wic}{\ensuremath{\texttt{WIC}}\xspace}

\title{Can LLMs Detect Their Confabulations?\\Estimating Reliability in Uncertainty-Aware Language Models}

\author{ {Tianyi Zhou}\\
	KTH Royal Institute of Technology\\
	Stockholm, Sweden \\
	\texttt{tzho@kth.se} \\
	\And
	{Johanne Medina} \\
	QCRI, HBKU\\
	Doha, Qatar\\
	\texttt{jomedina@hbku.edu.qa} \\
    \And
	{Sanjay Chawla} \\
	QCRI, HBKU\\
	Doha, Qatar\\
	\texttt{schawla@hbku.edu.qa} \\
}

\begin{document}

\maketitle

\begin{abstract}
Large Language Models (LLMs) are prone to generating fluent but incorrect content, known as confabulation, which poses increasing risks in multi-turn or agentic applications where outputs may be reused as context. In this work, we investigate how in-context information influences model behavior and whether LLMs can identify their unreliable responses. We propose a reliability estimation that leverages token-level uncertainty to guide the aggregation of internal model representations. Specifically, we compute aleatoric and epistemic uncertainty from output logits to identify salient tokens and aggregate their hidden states into compact representations for response-level reliability prediction. Through controlled experiments on open QA benchmarks, we find that correct in-context information improves both answer accuracy and model confidence, while misleading context often induces confidently incorrect responses, revealing a misalignment between uncertainty and correctness. Our probing-based method captures these shifts in model behavior and improves the detection of unreliable outputs across multiple open-source LLMs. These results underscore the limitations of direct uncertainty signals and highlight the potential of uncertainty-guided probing for reliability-aware generation.
\end{abstract}
\section{Introduction}
As large language models (LLMs) and generative AI tools become increasingly integrated into real-world applications, the need to quantify and interpret their uncertainty grows more urgent~\citep{sriramanan2024llm, Sensoy_Kaplan_Julier_Saleki_Cerutti_2025}. This is particularly important in multi-turn and agentic settings, where models operate autonomously and where contextual information (e.g. retrieved passages, prior conversation history, or agent-generated messages) plays a central role in shaping model behavior.

Should LLMs rely on their parametric, internalized knowledge or act as adaptive reasoning engines that synthesize and respond to external information? The growing adoption of Retrieval-Augmented Generation (RAG) pipelines and coordination protocols like the Model Context Protocol (MCP) highlights the urgency of understanding how context changes model behavior. 

\begin{figure}[]
\centering
\includegraphics[width=0.8\textwidth]{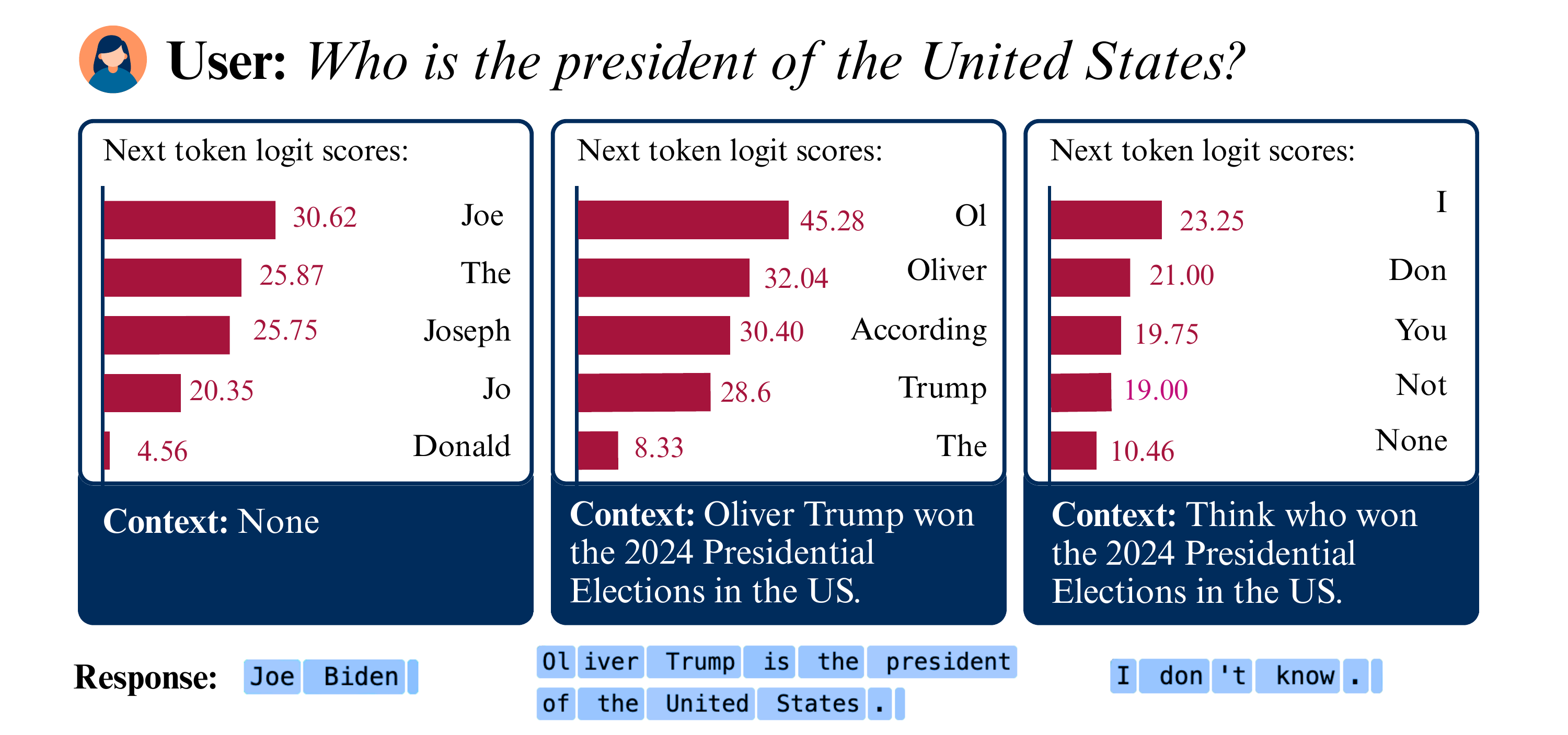}
\caption{Motivation example illustrating how next-token logit scores shift under varying context. Following evidential deep learning intuitions, we interpret logit values as token-level evidence. Without context, the model generates a correct but outdated answer with moderate logit scores. When exposed to misleading context, the model produces incorrect output with higher logit scores—indicating overconfidence. A neutral context leads to more distributed logits and a cautious response.}
\label{fig:motivation_example}
\end{figure}

When does external context enhance model reliability, and when does it induce new failure modes? 
Figure~\ref{fig:motivation_example} provides a motivating example. We prompt the model with the question \emph{“Who is the president of the United States?”} under three settings: no context, misleading context, and neutral context. In the absence of external information, \qwen answers \emph{``Joe Biden''}, a correct response at training time, although outdated. When presented with a misleading claim, e.g., \emph{``Oliver Trump won the 2024 Presidential Elections in the US''}, the model not only adopts this falsehood but does so with higher logit scores, which we interpret as stronger token-level evidence.
This behavior reflects a key insight from evidential deep learning \citep{sensoy2018evidentialdeeplearningquantify} where higher logits can be treated as higher evidence in favor of a particular prediction. The figure illustrates how in-context misinformation can affect the model’s internal evidence distribution, often leading to incorrect predictions made with high confidence. In contrast, a neutral prompt generates a more distributed and uncertain logit profile, resulting in a hedged response.


This observation motivates our first research question: \textit{How does in-context information influence model behavior and token-level uncertainty?} To investigate this, we design a controlled experimental framework in which the input query remains fixed while the surrounding context is systematically varied to either be omitted, accurate, or intentionally misleading. This controlled setup enables us to isolate the effect of contextual information on both the model’s output and its uncertainty profile. Our results indicate that accurate context generally improves response correctness and reduces uncertainty. In contrast, a misleading context often leads to confidently incorrect answers. This misalignment between confidence and correctness raises significant concerns for reliability, especially in retrieval-augmented and multi-agent settings where context is dynamically generated and potentially error-prone.

Having observed this limitation, we ask a second question: \textit{can internal signals, such as token-level uncertainty and hidden states, be used to detect when a model’s output is unreliable?}  
To investigate this, we develop probing-based classifiers that operate on token-level hidden representations, using uncertainty-guided token selection to form reliability features. We find that these classifiers consistently outperform direct uncertainty metrics and that aggregating features from high-uncertainty tokens leads to more accurate predictions of response correctness.

\textbf{This work makes three core contributions.} First, we present a context-controlled evaluation framework that reveals how LLMs transition between correct and incorrect responses depending on the quality of context. Second, we show that token-level uncertainty does not always align with correctness, particularly under misleading context, highlighting an underexplored vulnerability in model calibration. Third, we propose a probing-based approach for response reliability detection that leverages internal model activations and uncertainty-aware feature selection, outperforming standard baselines across tasks and models.

Our findings point to both the promise and limitations of using uncertainty as a signal for reliability in language models, and emphasize the importance of calibrating models not just at the output level, but also concerning the context they consume.

\section{Related Works}
Hallucinations are commonly categorized into \textit{factuality} errors, where outputs contradict reality, and \textit{faithfulness} errors, where responses diverge from provided context or instructions~\citep{qin2025dontlethallucinatepremise, Huang_Yu_Ma_Zhong_Feng_Wang_Chen_Peng_Feng_Qin_2025}. A particularly challenging subtype is \textit{confabulations}, which are fluent but ungrounded generations that may differ from the truth only subtly, making them difficult to detect~\citep{sui2024confabulationsurprisingvaluelarge, ji2023survey, reinhard2025fact}. \citet{orgad2024llms} further distinguish between cases where the model lacks relevant knowledge and those where it encodes the correct answer but fails to express it. These issues are compounded by overconfidence, where models assign high certainty to incorrect responses~\citep{li2024confidencemattersrevisitingintrinsic}, sometimes due to distribution shift leading to inflated confidence under unfamiliar inputs~\citep{wu2022distributioncalibrationoutofdomaindetection}. Understanding model uncertainty becomes crucial, as models should ideally respond with "I don't know" rather than hallucinating plausible-sounding but incorrect responses~\citep{ma2025estimating}.

\spara{Detection and Mitigation.}
Hallucination detection methods can be broadly divided into white-box and black-box approaches. White-box methods require access to model internals, leveraging probability signals, out-of-distribution cues, or hidden-state analysis, including techniques that locate where factual associations are stored~\citep{orgad2024llms}. 
Black-box methods rely solely on output text; they often generate multiple responses to assess consistency~\citep{Yadkori_Kuzborskij_Stutz_György_Fisch_Doucet_Beloshapka_Weng_Yang_Szepesvári_et2024}. Zero-shot techniques like SelfCheckGPT~\citep{manakul2023selfcheckgpt} evaluate internal agreement, while supervised detectors such as Lynx~\citep{lin2024lynx} leverage annotated data. Semantic-entropy methods capture meaning-level uncertainty~\citep{farquhar2024detecting, Yadkori_Kuzborskij_György_Szepesvári_2024}. Benchmarking resources like HaluBench~\citep{lin2024lynx} standardize evaluation, though context-dependent errors remain difficult. Beyond short-form settings, LongFact and SAFE~\citep{wei2024longformfactualitylargelanguage} assess long-form factuality by decomposing responses into claims and verifying them via search-augmented LLM agents. 
Similar to our approach, \citet{orgad2024llms} and \citet{obeso2025real} train classifiers on LLM internal representations of the exact answer token and associated entity tokens. Unlike their methods, ours requires no separate token-extraction pipeline for feature aggregation. 


Mitigation strategies include knowledge grounding via RAG~\citep{mallen2022trustworthy} and reasoning enhancement through chain-of-thought prompting (CoT)~\citep{wei2022chain}, though CoT may inadvertently increase confidence in incorrect outputs. Post-hoc verification methods such as Chain-of-Verification (CoVe)~\citep{dhuliawala2023chain} refine responses but increase inference cost. More recently, SLED~\citep{zhang2025sledselflogitsevolution} improves factual accuracy without external retrieval or fine-tuning by contrasting early- and late-layer logits and using these signals for self-correction. 

\spara{Uncertainty and Calibration.}
LLMs are frequently miscalibrated, producing incorrect answers with unwarranted confidence~\citep{Abdar_Pourpanah_Hussain_Rezazadegan_Liu_Ghavamzadeh_Fieguth_Cao_Khosravi_Acharya_2021}. Approaches such as self-consistency decoding~\citep{wang2023selfconsistency} aim to align confidence with correctness better but remain sensitive to prompt formulation and decoding variability. While in-context learning (ICL) enables rapid generalization, it also introduces reliability risks: misleading prompts or poorly chosen examples can induce hallucinations or biased outputs~\citep{simhi2024constructing,an2023skill}. Current models lack mechanisms to validate or reject flawed contextual signals, motivating uncertainty-aware generation resilient to noisy or adversarial context. Closely aligned, work in risk-aware classification formalizes how predictive uncertainty should guide decision-making; \citet{Sensoy_Kaplan_Julier_Saleki_Cerutti_2025} extend evidential deep learning to support abstention under high epistemic uncertainty.

\section{Preliminary}
\label{sec:preliminary}
We begin by introducing key notations and definitions that will be used throughout the paper.

\spara{Generation process.}
Let $\mathcal{M}$ be a pre-trained language model with tokenizer vocabulary $\mathcal{V} = \{\tau_1, \tau_2, \ldots, \tau_{|V|}\}$. Given a user-specified question $q$, the tokenizer encodes it into a prompt vector $\v+p = (p_1, \ldots, p_n)$, which is used by $\mathcal{M}$ to autoregressively generate a response vector $\v+y = (y_1, \ldots, y_T)$. At each generation step $t$, the model outputs logits $\v+a_t \in \mathbb{R}^{|\mathcal{V}|}$, which are converted to a probability distribution over $\mathcal{V}$ via the softmax function. A token $y_t$ is then sampled according to a decoding strategy:
\begin{equation}
y_t \sim P_{\mathcal{M}}(\mathcal{V} \mid \v+p, \v+y_{<t}),
\end{equation}
where $\v+y_{<t} = (y_1, \ldots, y_{t-1})$.

The generation continues token by token until a special end-of-sequence token $\texttt{[EOS]} \in \mathcal{V}$ is produced. The overall generation process can be deterministic:
\begin{equation}
\v+y = \arg\max_{y_1, \ldots, y_T} \prod_{t=1}^{T} P_{\mathcal{M}}(y_t \mid \v+p, \v+y_{<t}),
\end{equation}
or stochastic, using methods such as top-$p$ sampling.

\spara{Uncertainty estimation.}
We estimate token-level uncertainty using the output logits of the model, following the Dirichlet-based framework of \citet{ma2025estimating, sensoy2018evidentialdeeplearningquantify}. Given the logits vector $\v+a_t$ at generation step $t$, we select the top-$K$ logits corresponding to the tokens with highest predicted values to construct a Dirichlet distribution. Let $\tau_k$ denote the token with the $k$-th highest logit, and define:
\begin{equation}
a_k = \mathcal{M}(\tau_k \mid \v+q, y_{<t}), \quad a_0 = \sum_{k=1}^{K} a_k,
\end{equation}
where $a_k$ serves as the evidence for token $\tau_k$, and $a_0$ is the total evidence.

The \emph{aleatoric uncertainty} (AU), capturing uncertainty from inherent data ambiguity, is defined as the expected entropy of the Dirichlet-distributed categorical distribution:
\begin{equation}
\mathrm{AU}(\v+a_t) = -\sum_{k=1}^{K} \frac{a_k}{a_0} \left( \psi(a_k + 1) - \psi(a_0 + 1) \right),
\label{eq:au}
\end{equation}
where $\psi(\cdot)$ denotes the digamma function.

The \emph{epistemic uncertainty} (EU), reflecting the model's confidence based on available evidence, is defined as:
\begin{equation}
\mathrm{EU}(\v+a_t) = \frac{K}{\sum_{k=1}^{K} (a_k + 1)}.
\label{eq:eu}
\end{equation}

In addition to the final-layer logits $\v+a_t$, LLMs produce internal representation vectors at each layer for every token. Let $\mathbf{h}^{(l)}_t \in \mathbb{R}^d$ denote the hidden state of the $t$-th token $y_t$ at layer $l$, where $d$ is the hidden dimension. For a generated response sequence $\mathbf{y} = (y_1, \ldots, y_T)$ of length $T$, the hidden states at layer $l$ form a matrix $\mathbf{H}^{(l)} = [\mathbf{h}^{(l)}_1, \ldots, \mathbf{h}^{(l)}_T] \in \mathbb{R}^{d \times T}$. These hidden states encode intermediate representations of the sequence, capturing progressively refined semantic and syntactic information across layers.


\spara{Model behavior.}  
When LLMs generate multiple responses to a given prompt, they may produce confabulations due to insufficient knowledge. We quantify this behavior by measuring the confabulation rate over $m$ sampled responses.

For each prompt $\v+p$, assume a ground-truth response vector $\v+y^{\star}$. Let $z \in \{0,1\}$ be a binary correctness label indicating whether a generated response is semantically correct. Specifically, we define a similarity function $S: \mathcal{Y} \times \mathcal{Y} \rightarrow \mathbb{R}$ that measures semantic similarity between two responses $\v+y, \v+y^{\star} \in \mathcal{Y}$. A response is considered correct if $S(\v+y, \v+y^{\star}) > \theta$, where $\theta$ is a predefined similarity threshold; that is,
\[
z = \begin{cases}
1, & \text{if } S(\v+y, \v+y^{\star}) > \theta, \\
0, & \text{otherwise}.
\end{cases}
\]

We then sample $M$ responses $\m+Y = (\v+y_1, \ldots, \v+y_M)$ for each prompt, and obtain the corresponding correctness vector $\v+z = (z_1, \ldots, z_M)$. The \emph{correctness ratio} $r \in [0,1]$ is defined as the fraction of correct responses:
\[
r = \frac{1}{M} \sum_{i=1}^M z_i.
\]

This ratio serves as an empirical proxy for the model's confidence: a high value implies that the model consistently produces correct responses, suggesting it has internalized the required knowledge; a low value suggests a lack of understanding or memorization.

To further categorize model behavior, we define two response regimes: \emph{mostly correct} (\texttt{C}), where $r > \tau_C$, and \emph{mostly wrong} (\texttt{E}), where $r < \tau_E$, with $\tau_C$ and $\tau_E$ being predefined thresholds.

\spara{In-context learning.}  
In addition to the prompt $\v+p$, LLMs can incorporate \emph{in-context information} during generation, such as demonstrations or retrieved passages, prepended to the input. This mechanism, known as \emph{in-context learning} (ICL), allows the model to adapt its output distribution at inference time without parameter updates. We investigate how the model's behavior and uncertainty change across different context settings, which is particularly relevant in agentic or multi-turn scenarios, where a model’s own outputs may be used as context in subsequent interactions.

Specifically, we define three context settings: no context (\texttt{WOC}), correct context (\texttt{WCC}), and incorrect or misleading context (\texttt{WIC}). Let $\mathcal{C} = \{\texttt{WCC}, \texttt{WIC}\}$ denote the set of context types involving additional input. For a given prompt, we compare the model’s error type across different context settings and define a subset of \emph{error-shifting questions}—those for which the model transitions between regimes (e.g., \texttt{WOC:C}~$\rightarrow$~\texttt{WIC:E}). This enables us to isolate instances where in-context information significantly alters the model’s response's correctness and uncertainty.

\spara{Research questions.} Having introduced our setup, we now introduce our research questions.
\begin{description}
    \item[RQ1:] \emph{How does in-context information influence model behavior and response uncertainty?}  
    We aim to quantify how the presence of correct or misleading context affects both the correctness of generated responses and the model’s confidence, as captured by uncertainty measures.

    \item[RQ2:] \emph{Can uncertainty signals be used to predict response reliability?}  
    We investigate whether epistemic and aleatoric uncertainty scores can serve as effective features for detecting whether a model’s response is factually reliable, and how these signals compare to other baselines.
\end{description}
In the following, we experimentally answer all these questions in detail. 

\section{The Influence of In-context Learning on Model Behavior and Uncertainty}


Large language models exhibit varying behaviors depending on the presence and quality of contextual information. In this section, we address \textbf{RQ1}: \emph{How does in-context information influence model behavior and response uncertainty?} 

By systematically comparing model outputs across different context conditions—no context, correct context, and misleading context—we aim to isolate the effect of external information on both model predictions and confidence. This setup enables a fine-grained analysis of how context modulates output correctness and how such changes are reflected in the distribution of uncertainty scores.

\spara{Experiment setup.}  
We design a controlled experiment using two benchmark QA datasets that include supporting passages: \hotpot~\citep{Yang0ZBCSM18} and \NQ~\citep{NQgoogle}. Both datasets provide ground-truth factual context, but do not include incorrect or misleading information. To evaluate model behavior under misleading conditions, we construct a smaller evaluation set by sampling 2{,}000 examples from \hotpot and 1{,}000 from \NQ, and use ChatGPT-4.1-mini to automatically rewrite the original supporting passages to introduce plausible but incorrect content.

We evaluate three large language models (LLMs): \fanar, \gemma, and \qwen. \fanar is an Arabic-centric LLM designed for multilingual understanding~\citep{fanarteam2025fanararabiccentricmultimodalgenerative}; \gemma is a publicly released instruction-tuned model by Google; and \qwen is a state-of-the-art bilingual (English-Chinese) model developed by Alibaba's DAMO Academy.

Next, we quantify the model response behavior on the questions $Q$. For each question prompt $\v+p_i$, we sample $15$ responses 
using stochastic decoding under each of the three context settings: without context (\woc), with correct context (\wcc), and with incorrect context (\wic). Each response $\v+y_{i}^{(j)}$ is labeled using GPT-4.1 mini, guided by a prompt to assess semantic equivalence with the ground truth answer. Based on these labels, we compute the correctness ratio and classify each prompt-response pair into response regimes. We set the correctness thresholds as $\tau_C > 0.6 $ and $\tau_E < 0.4$.
For detailed implementations, see Appendix \ref{ap:implementation-details}.

\begin{figure}[ht]
\centering
\includegraphics[width=0.48\textwidth]{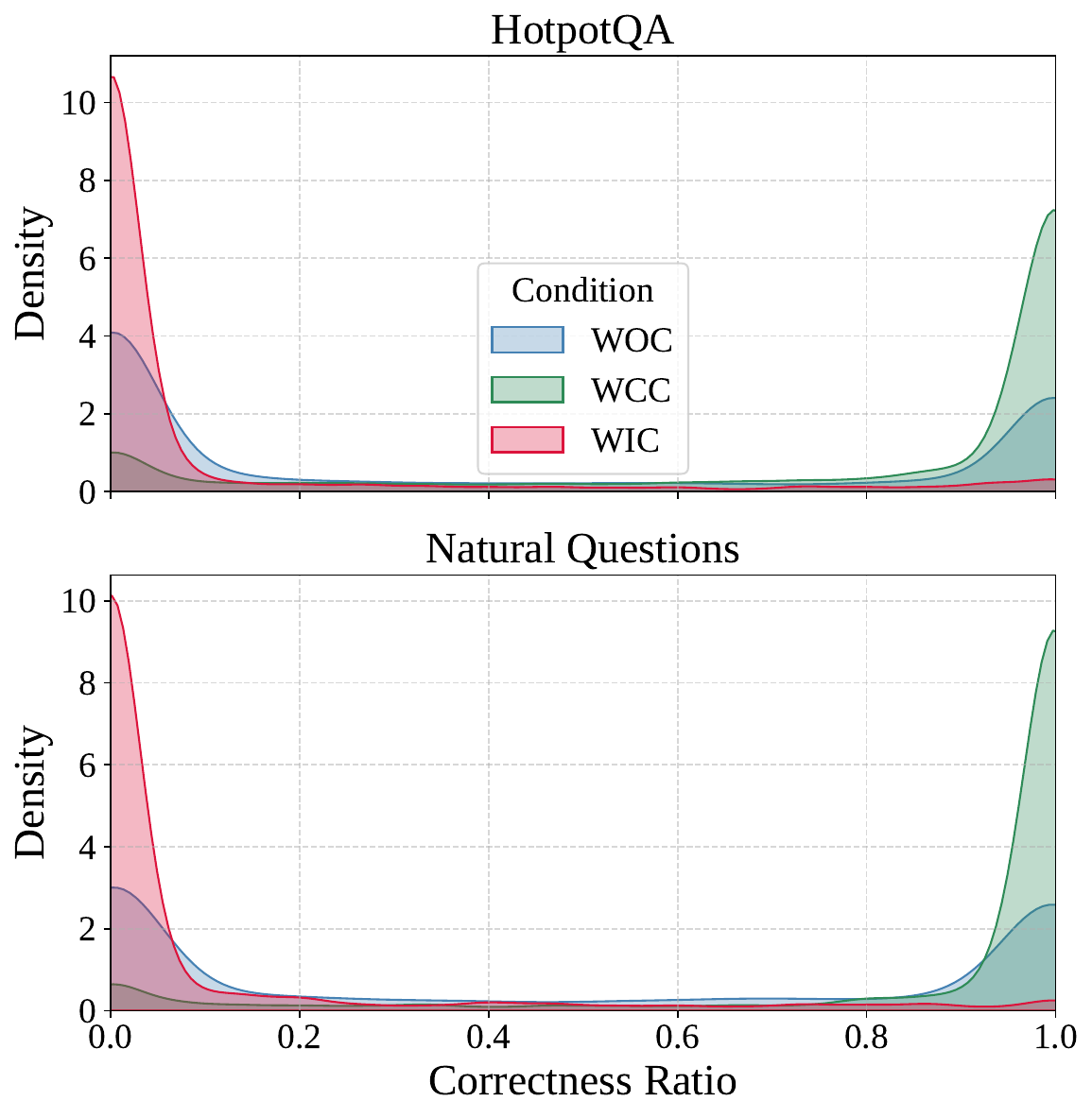}
\caption{Impact of contextual information on response correctness.
Distribution of aggregated correctness ratios on the HotpotQA and Natural Questions datasets  across three context conditions:  without context (\texttt{WOC}), correct context (\texttt{WCC}), and incorrect context (\texttt{WIC}).
}
\label{fig:behavior-transition}
\end{figure}
\spara{Effect of context on correctness ratio.}  Figure~\ref{fig:behavior-transition} illustrates the distribution of correctness ratios for questions under three context conditions: no context (\texttt{WOC}), correct context (\texttt{WCC}), and incorrect context (\texttt{WIC}), across the \hotpot and \NQ datasets. The correctness ratio reflects the fraction of generated responses labeled as semantically correct out of $K$ samples per question.

We observe a clear shift in distributions when context is introduced. Providing correct context (\texttt{WCC}) significantly increases the proportion of high correctness ratios (peaking near 1.0), suggesting that access to relevant external information enhances model reliability. In contrast, introducing incorrect or misleading context (\texttt{WIC}) leads to a pronounced concentration near zero, indicating that models often produce consistently wrong responses with misleading input. The baseline (\texttt{WOC}) condition sits between these two extremes, showing a more dispersed distribution.

These patterns confirm that context strongly modulates model behavior. Accurate context improves consistency and correctness, while misleading context systematically degrades performance. This highlights the importance of validating contextual inputs, especially in multi-turn or retrieval-augmented generation settings.

\begin{figure*}[ht]
\centering
\includegraphics[width=\textwidth]{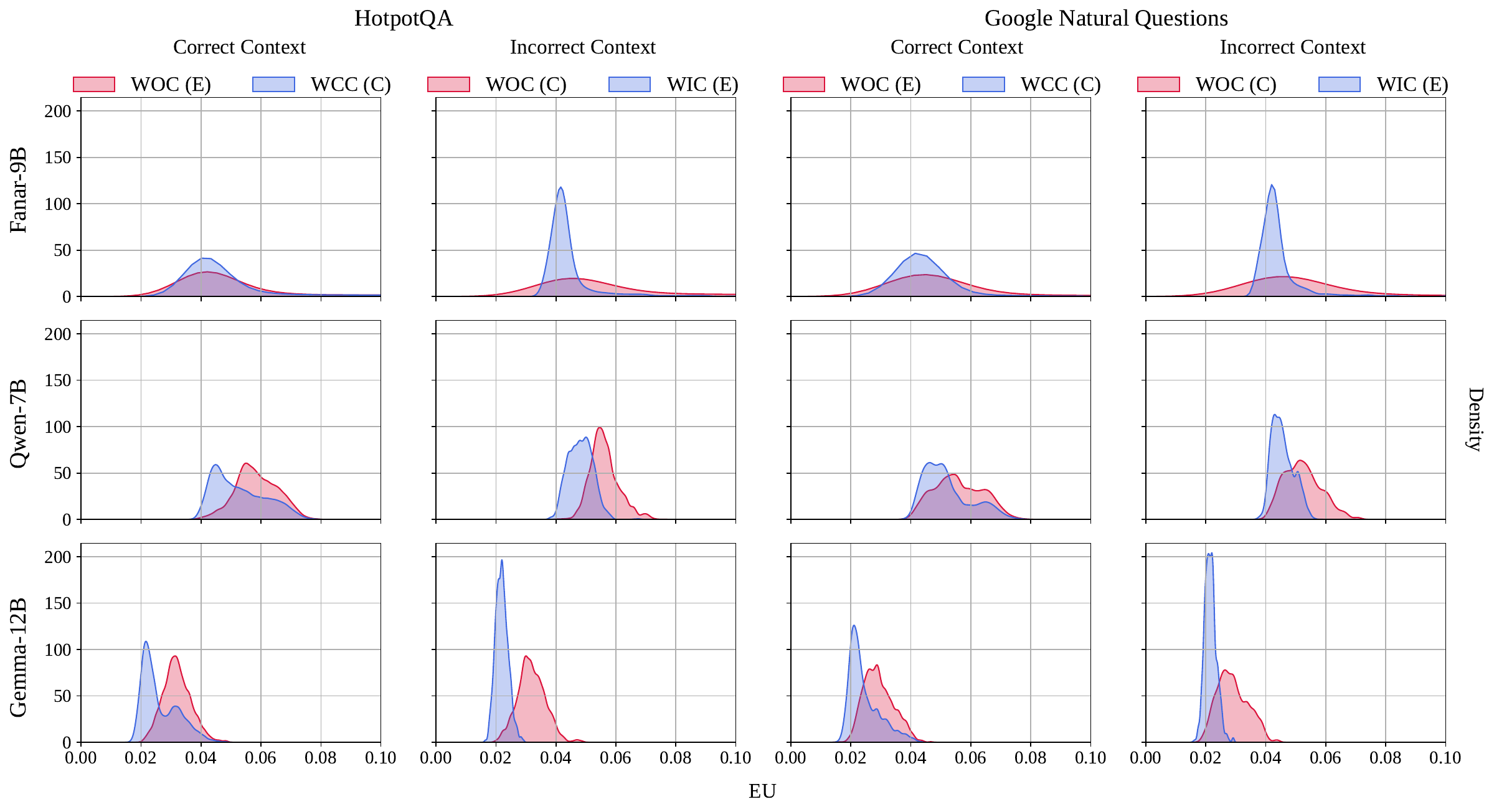}
\caption{Model behavior transitions and epistemic uncertainty (EU) distribution shifts across \hotpot and \NQ for three models (\fanar, \qwen, \gemma). Each subplot displays the distribution of lower-bound epistemic uncertainty scores for subsets of questions whose correctness regime changes between the no-context (\texttt{WOC}) and context-enhanced (\texttt{WCC} or \texttt{WIC}) settings.
We focus on two key transitions: (1) \texttt{WOC:E}~$\rightarrow$~\texttt{WCC:C}, where injecting correct context into previously incorrect responses leads to improved correctness and decreased uncertainty; and (2) \texttt{WOC:C}~$\rightarrow$~\texttt{WIC:E}, where misleading context causes the model to produce incorrect responses with sustained low uncertainty. These shifts highlight how in-context information modulates both model predictions and confidence, revealing risks of overconfident confabulations in the presence of incorrect input.}
\label{fig:eu_shift}
\end{figure*}

\spara{Uncertainty profiles of different response regimes.}  
To understand the uncertainty characteristics of responses within specific behavioral regimes, we analyze the \emph{uncertainty region} of each generated response. Specifically, we define the \emph{lower bound} of uncertainty as the average of the $K$ smallest token-level uncertainty scores, and the \emph{upper bound} as the average of the $K$ largest scores. These bounds capture the most confident and most uncertain regions of the response, respectively. We focus our analysis on subsets of questions $Q'$ that exhibit a transition in response regime under different context conditions (e.g., from mostly incorrect to mostly correct). Specifically, we focus on two key behavior transitions:
\begin{itemize}
    \item \texttt{WOC:E}~$\rightarrow$~\texttt{WCC:C}: Questions initially classified as mostly wrong (\texttt{E}) without context become mostly correct (\texttt{C}) with correct context. This indicates the model lacks sufficient parametric knowledge but can utilize external information when provided.
    
    \item \texttt{WOC:C}~$\rightarrow$~\texttt{WIC:E}: Questions initially mostly correct (\texttt{C}) degrade to mostly wrong (\texttt{E}) when given misleading context. This highlights the model's vulnerability to confabulations triggered by incorrect external information, despite possessing sufficient internal knowledge.
\end{itemize}


Figure~\ref{fig:eu_shift} visualizes the distribution of lower-bound epistemic uncertainty across these subsets using kernel density estimation (KDE), allowing for comparison of uncertainty profiles before and after the context shift. Results are shown for three models—\fanar, \qwen, and \gemma—on the \hotpot and \NQ datasets. For completeness, we also replicate this analysis on the \NQ dataset with the newly released \oss by OpenAI under the same experimental settings, with results shown in Figure~\ref{fig:oss_nq} in Appendix~\ref{ap:additional-experimental-results}.

\emph{Correct context reduces uncertainty.} 
As expected, we observe a clear and consistent decrease in epistemic uncertainty in the transition from incorrect responses without context to correct responses with context (\texttt{WOC:E}$\rightarrow$\texttt{WCC:C}). Across all models, the KDE curves corresponding to the \texttt{WCC:C} setting shift leftward relative to those from the \texttt{WOC:E} setting, indicating that providing accurate contextual information not only improves answer correctness but also increases model confidence.
This effect is particularly pronounced for \qwen and \gemma, where the uncertainty distributions in the \texttt{WCC:C} condition are sharply concentrated around low epistemic uncertainty values.

\emph{Misleading context induces confident errors.}  
We analyze the setting where models transition from correct predictions without context (\texttt{WOC:C}) to incorrect predictions with misleading context (\texttt{WIC:E}). Ideally, such a transition should result in higher epistemic uncertainty, reflecting the model’s recognition of ambiguity or conflict, visualized as broader, right-shifted distributions. However, all models instead show a contraction in their EU distributions, with \texttt{WIC:E} responses exhibiting sharper and more left-skewed profiles.

\fanar, despite appearing flat under correct context conditions, exhibits a notable increase in peakedness and reduced variance under misleading context, indicating an unjustified confidence in its wrong answers. This suggests that Fanar is responsive to misleading context and exhibits similar calibration issues as the other models, even if the mean EU shift is modest.
\qwen also produces more confident predictions under misleading context, with \texttt{WIC:E} curves shifting left and becoming narrower relative to \texttt{WOC:C}.
\gemma shows the most extreme behavior, with the narrowest and most left-shifted \texttt{WIC:E} distribution. This reflects strong contextual dependence but very poor calibration when that context is misleading.


These results reveal a dual role of contextual information in large language model behavior. When context is accurate, it reliably improves both correctness and model confidence. However, misleading context can cause models to produce incorrect answers with high certainty. These findings align with expectations and emphasize the importance of robust uncertainty estimation in detecting context-induced confabulations. They motivate future research in reliability-aware generation and mechanisms for validating or filtering context in multi-turn or retrieval-augmented generation settings. In the following section, we investigate how to use uncertainty information to guide the response reliability detection. 

\section{Effectiveness of Uncertainty-Guided Probing for Reliability Detection}
As shown in our analysis of \textbf{RQ1}, token-level uncertainty is not always aligned with correctness, particularly under in-context learning. In the presence of misleading information, models may produce confident yet incorrect responses—a phenomenon that raises concerns in multi-turn or retrieval-augmented settings, where such confabulated outputs may be reused as context in future turns. This observation underscores the limitations of using uncertainty alone as a reliability signal when external context is present.

However, in scenarios where the model relies solely on its internal parameters (i.e., without additional context), uncertainty may still provide meaningful cues about response reliability. This motivates our investigation in \textbf{RQ2}: \emph{Can token-level uncertainty, when combined with internal representations, be used to detect unreliable responses?} 

We explore this question by training probing classifiers on token-level hidden states from various layers and positions, using both static and uncertainty-aware token selection strategies. Our goal is to assess whether internal signals, especially those grounded in model confidence, can serve as reliable indicators of output correctness.

\begin{figure*}[ht]
\centering
\includegraphics[width=\textwidth]{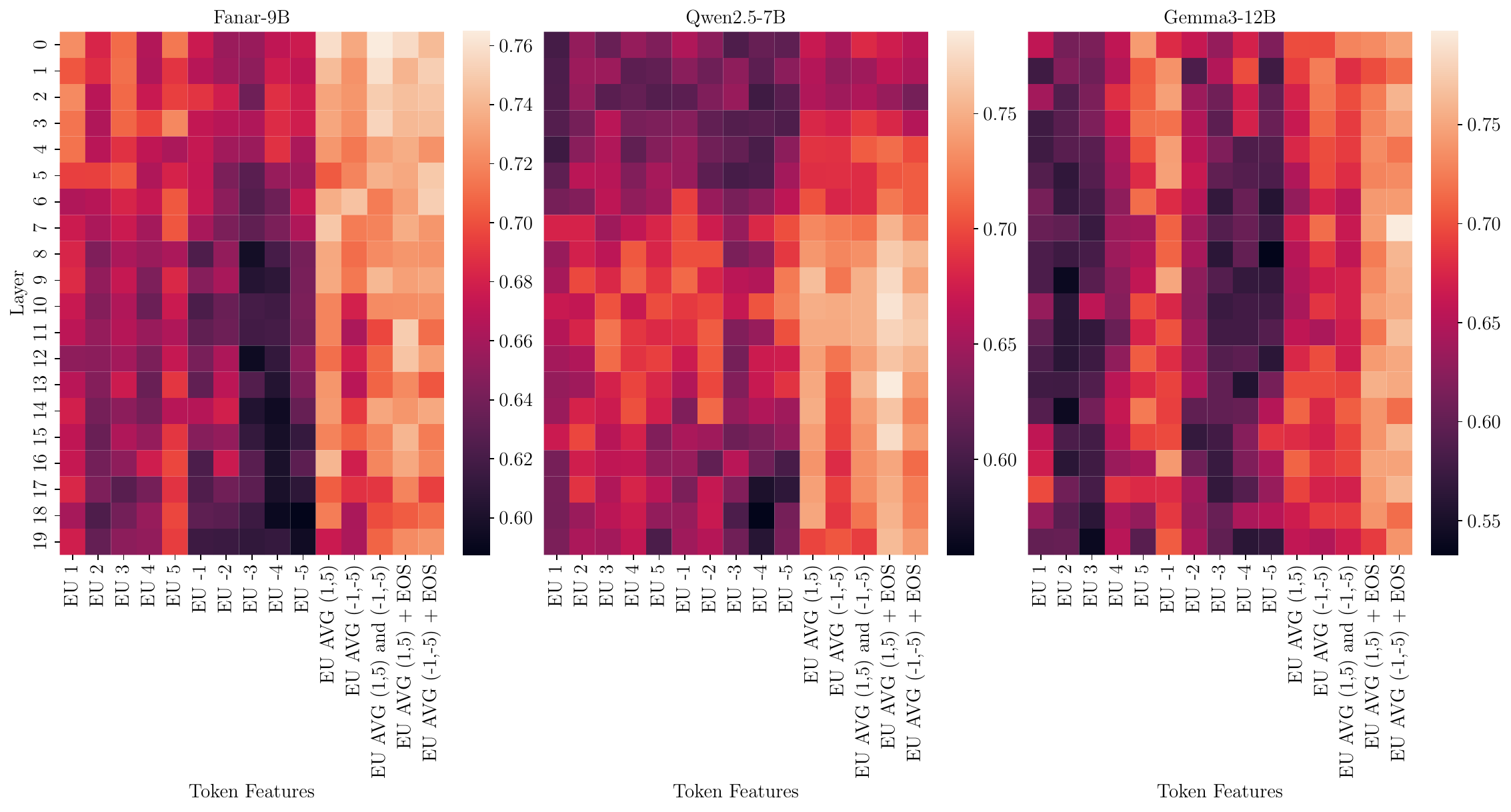}
\caption{AUROC scores of probing classifiers across the last 20 layers using different token-level features, evaluated on the \triviaqa dataset for \fanar, \qwen, and \gemma. From left to right, columns correspond to probing with single tokens ranked by epistemic uncertainty: the $k$ smallest (EU 1 to EU 5) and the $k$ largest (EU -1 to EU -5). Aggregated features (EU AVG) formed by averaging hidden states across selected tokens yield the highest detection performance across all models.}
\label{fig:token_feature_heatmap}
\end{figure*}

\spara{Response reliability detection.} We consider the following method from the related literature of uncertainty, reliability, and hallucination detection.
\begin{itemize}
    \item \textbf{LogProb:} This method computes the mean of log-probability scores of the generated tokens \cite{Yadkori_Kuzborskij_György_Szepesvári_2024}
        \begin{equation*}
             \frac{1}{T} \sum_{t=1}^{T} \log \mathbb{P}(y_t|\v+p,\v+y_{<t}) .
        \end{equation*}
       
    \item \textbf{P(True):} This method prompts the LLMs to judge whether their answer is correct. Our prompt followed the following template from \citet{kadavath2022language}.

    \item \textbf{LogTokU:} This method computes the aggregated aleatoric and epistemic uncertainty to predict the response reliability. We follow the aggregation method from \citet{ma2025estimating}.
    
    \item \textbf{Probing.}  
    We train lightweight classifiers on token-level hidden states $\mathbf{h}^{(l)}_t$ to predict response-level reliability following previous work \cite{DBLP:conf/nips/0002PVPW23}. We consider several token selection strategies:
    \begin{itemize}
        \item \textbf{Probe(EOS):} Uses the final generated token $\mathbf{h}^{(l)}_{T}$.
        \item \textbf{Probe(Exact):} Selects tokens aligned with the exact answer span~\citep{orgad2024llms}.
        \item \textbf{Probe(EU):} Selects the single token with either the highest or lowest epistemic uncertainty score.
        \item \textbf{Probe(AVG):} Average hidden states across selected token subsets (e.g., top-$k$ uncertain tokens or fixed positions) to form an aggregated feature representation.
    \end{itemize}
    \end{itemize}
    
\spara{Performance metric.} We use the area under the receiver operating characteristic curve (AUROC) to evaluate the performance of reliability detectors. This metric summarizes the model's ability to distinguish between positive and negative cases across all classification thresholds, effectively balancing sensitivity (true positive rate) and specificity (false positive rate).

\begin{table}[ht]
    \centering
    \small
    \begin{tabular}{llccc}
        \toprule
        \textbf{Model} & \textbf{Method} & \textbf{TruthfulQA} & \textbf{TriviaQA} & \textbf{Math} \\
        \midrule
        \multirow{7}{*}{\textbf{Fanar}}
            & \textbf{LogProb}          & 0.597 & \underline{0.774} & 0.757 \\
            & \textbf{P(true)}          & 0.530 & 0.672 & 0.635 \\
            & \textbf{LogTokU}          & 0.541 & 0.683 & 0.666 \\
            & \textbf{Prob(Exact)}      & \underline{0.711} & \textbf{0.783} & \underline{0.827} \\
            & \textbf{Probe(EOS)}       & 0.706 & 0.739 & 0.790 \\
            & \textbf{Probe(EU)}        & 0.709 & 0.751 & 0.794\\
            & \textbf{Probe(AVG)}       & \textbf{0.734} & 0.765 & \textbf{0.833} \\
        \midrule
        \multirow{7}{*}{\textbf{Qwen}}
            & \textbf{LogProb}          & 0.591 & 0.774 & 0.635 \\
            & \textbf{P(true)}          & 0.537 & 0.736 & 0.664 \\
            & \textbf{LogTokU}          & 0.642 & 0.773 & 0.565 \\
            & \textbf{Prob(Exact)}      & 0.758 & 0.781 & 0.627 \\
            & \textbf{Probe(EOS)}       & \textbf{0.794} & \textbf{0.812} & \textbf{0.703} \\
            & \textbf{Probe(EU)}        & 0.759 & 0.754 & 0.646 \\
            & \textbf{Probe(AVG)}       & \underline{0.761} & \underline{0.786} & \underline{0.699} \\
        \midrule
        \multirow{7}{*}{\textbf{Gemma}}
            & \textbf{LogProb}          & 0.545 & 0.806 & 0.683 \\
            & \textbf{P(true)}          & 0.598 & 0.631 & 0.779 \\
            & \textbf{LogTokU}          & 0.489 & 0.611 & \underline{0.791} \\
            & \textbf{Prob(Exact)}      & 0.728 & 0.796 & 0.773 \\
            & \textbf{Probe(EOS)}       & \underline{0.728} & \underline{0.810} & \textbf{0.834} \\
            & \textbf{Probe(EU)}        & 0.687 & 0.751 & 0.669 \\
            & \textbf{Probe(AVG)}       & \textbf{0.733} & \textbf{0.818} & 0.786 \\
        \bottomrule
    \end{tabular}
    \caption{Comparison of probing methods across \fanar, \qwen, and \gemma models on three datasets. We report AUROC scores (3-decimal precision). Bold indicates the best in each column; underlined indicates the second-best.}
        \label{tab:baselines}
\end{table}

\spara{Reliability detection cross layers and tokens.}
Figure~\ref{fig:token_feature_heatmap} presents the AUROC scores of probing classifiers trained on hidden states from the last 20 layers, using different token-level feature strategies under the epistemic uncertainty setup. Each heatmap column represents a token selection method ranging from single-token probing (e.g., using the token with highest or lowest uncertainty) to aggregated representations computed by averaging hidden states across multiple tokens.

We observe that individual token features (left columns) often yield weaker performance, especially in earlier layers. In contrast, aggregated features (right columns) consistently lead to better classification results. This trend holds across all models. In particular, strategies like \texttt{EU AVG (1-5) + EOS} achieve the highest AUROC scores, especially when features are extracted from middle to upper layers. These results suggest that combining multiple token-level signals enhances the robustness of response-level reliability detection.

\spara{Comparison with Uncertainty-Based Baselines.}
Next, we compare the reliability detection performance of different methods. Table~\ref{tab:baselines} summarizes the AUROC performance of different methods across three LLMs (\fanar, \qwen, \gemma) and three QA datasets (\truthfulqa, \triviaqa, \mathqa). Probing methods clearly outperform uncertainty-only baselines such as \textbf{LogProb} and \textbf{P(true)}, demonstrating the added value of internal model representations. Among all methods, \textbf{Probe(AVG)} yields the best overall performance, followed by \textbf{Probe(EOS)} and \textbf{Probe(EU)}. Although \gemma achieves strong performance with \textbf{LogTokU} on \truthfulqa, probing methods are more robust across tasks. Notably, performance is higher on \triviaqa and \mathqa, indicating that response reliability is more predictable in factoid-style and structured QA than in open-ended questions.

These findings highlight the effectiveness of token-level probing for reliability detection. Aggregating hidden states over uncertain or boundary tokens provides a strong signal and consistently outperforms uncertainty-only baselines. This supports the utility of internal representations in enabling more reliable LLM-generated outputs.

\section{Conclusion and Future Work}
In this work, we investigate how large language models respond to different types of contextual input, with a focus on identifying and understanding failure modes. We found that providing accurate context improves both model accuracy and confidence, whereas misleading context can lead to confidently incorrect outputs. This reveals a misalignment between uncertainty estimates and actual correctness, particularly under in-context learning, and raises concerns about confabulated responses being reused in multi-turn or retrieval-augmented generation. To better understand and potentially identify unreliable responses, we explored a probing-based approach that leverages token-level hidden states and uncertainty-guided token selection. Our experiments across multiple models and datasets suggest that this approach offers improved performance over direct uncertainty-based baselines. In particular, aggregating features from multiple tokens—especially those with high uncertainty—provides more informative signals for predicting response reliability.

While our analysis focuses on question answering tasks, extending these techniques to open-ended generation and multi-turn dialogue remains an open challenge. Future work could explore incorporating reliability signals into generation-time decisions, combining probing-based methods with retrieval validation, and developing safeguards to limit the propagation of confabulated content in interactive applications.


\begin{thebibliography}{32}
\providecommand{\natexlab}[1]{#1}
\providecommand{\url}[1]{\texttt{#1}}
\expandafter\ifx\csname urlstyle\endcsname\relax
  \providecommand{\doi}[1]{doi: #1}\else
  \providecommand{\doi}{doi: \begingroup \urlstyle{rm}\Url}\fi

\bibitem[Abdar et~al.(2021)Abdar, Pourpanah, Hussain, Rezazadegan, Liu, Ghavamzadeh, Fieguth, Cao, Khosravi, Acharya, Makarenkov, and Nahavandi]{Abdar_Pourpanah_Hussain_Rezazadegan_Liu_Ghavamzadeh_Fieguth_Cao_Khosravi_Acharya_2021}
Moloud Abdar, Farhad Pourpanah, Sadiq Hussain, Dana Rezazadegan, Li~Liu, Mohammad Ghavamzadeh, Paul Fieguth, Xiaochun Cao, Abbas Khosravi, U.~Rajendra Acharya, Vladimir Makarenkov, and Saeid Nahavandi.
\newblock A review of uncertainty quantification in deep learning: Techniques, applications and challenges.
\newblock \emph{Inf. Fusion}, 76\penalty0 (C):\penalty0 243–297, December 2021.
\newblock ISSN 1566-2535.
\newblock \doi{10.1016/j.inffus.2021.05.008}.
\newblock URL \url{https://doi.org/10.1016/j.inffus.2021.05.008}.

\bibitem[An et~al.(2023)An, Zhang, Wang, Yuan, Li, and Yih]{an2023skill}
Shuwen An, Zihang Zhang, Zihua Wang, Han Yuan, Xiang~Lisa Li, and Wen-tau Yih.
\newblock Skill-based few-shot prompting for large language models.
\newblock In \emph{Proceedings of the 2023 Conference on Empirical Methods in Natural Language Processing}, 2023.

\bibitem[Dhuliawala et~al.(2024)Dhuliawala, Komeili, Xu, Raileanu, Li, Celikyilmaz, and Weston]{dhuliawala2023chain}
Shehzaad Dhuliawala, Mojtaba Komeili, Jing Xu, Roberta Raileanu, Xian Li, Asli Celikyilmaz, and Jason Weston.
\newblock Chain-of-verification reduces hallucination in large language models.
\newblock In Lun-Wei Ku, Andre Martins, and Vivek Srikumar (eds.), \emph{Findings of the Association for Computational Linguistics: ACL 2024}, pp.\  3563--3578, Bangkok, Thailand, August 2024. Association for Computational Linguistics.
\newblock \doi{10.18653/v1/2024.findings-acl.212}.
\newblock URL \url{https://aclanthology.org/2024.findings-acl.212/}.

\bibitem[Farquhar et~al.(2024)Farquhar, Kossen, Kuhn, and Gal]{farquhar2024detecting}
Sebastian Farquhar, Jannik Kossen, Lorenz Kuhn, and Yarin Gal.
\newblock Detecting hallucinations in large language models using semantic entropy.
\newblock \emph{Nature}, 630\penalty0 (8017):\penalty0 625--630, 2024.

\bibitem[Huang et~al.(2025)Huang, Yu, Ma, Zhong, Feng, Wang, Chen, Peng, Feng, Qin, and Liu]{Huang_Yu_Ma_Zhong_Feng_Wang_Chen_Peng_Feng_Qin_2025}
Lei Huang, Weijiang Yu, Weitao Ma, Weihong Zhong, Zhangyin Feng, Haotian Wang, Qianglong Chen, Weihua Peng, Xiaocheng Feng, Bing Qin, and Ting Liu.
\newblock A survey on hallucination in large language models: Principles, taxonomy, challenges, and open questions.
\newblock \emph{ACM Transactions on Information Systems}, 43\penalty0 (2):\penalty0 1–55, March 2025.
\newblock ISSN 1046-8188, 1558-2868.
\newblock \doi{10.1145/3703155}.

\bibitem[Ji et~al.(2023)Ji, Lee, Frieske, Yu, Su, Xu, Ishii, Bang, Madotto, and Fung]{ji2023survey}
Ziwei Ji, Nayeon Lee, Rita Frieske, Tiezheng Yu, Dan Su, Yan Xu, Etsuko Ishii, Ye~Jin Bang, Andrea Madotto, and Pascale Fung.
\newblock Survey of hallucination in natural language generation.
\newblock \emph{ACM Computing Surveys}, 55\penalty0 (12), March 2023.
\newblock ISSN 0360-0300.
\newblock \doi{10.1145/3571730}.
\newblock URL \url{https://doi.org/10.1145/3571730}.

\bibitem[Kadavath et~al.(2022)Kadavath, Conerly, Askell, Henighan, Drain, Perez, Schiefer, Hatfield-Dodds, DasSarma, Tran-Johnson, et~al.]{kadavath2022language}
Saurav Kadavath, Tom Conerly, Amanda Askell, Tom Henighan, Dawn Drain, Ethan Perez, Nicholas Schiefer, Zac Hatfield-Dodds, Nova DasSarma, Eli Tran-Johnson, et~al.
\newblock Language models (mostly) know what they know.
\newblock \emph{arXiv preprint arXiv:2207.05221}, 2022.

\bibitem[Kwiatkowski et~al.(2019)Kwiatkowski, Palomaki, Redfield, Collins, Parikh, Alberti, Epstein, Polosukhin, Devlin, Lee, et~al.]{NQgoogle}
Tom Kwiatkowski, Jennimaria Palomaki, Olivia Redfield, Michael Collins, Ankur Parikh, Chris Alberti, Danielle Epstein, Illia Polosukhin, Jacob Devlin, Kenton Lee, et~al.
\newblock Natural questions: a benchmark for question answering research.
\newblock \emph{Transactions of the Association for Computational Linguistics}, 7:\penalty0 453--466, 2019.

\bibitem[Li et~al.(2023)Li, Patel, Vi{\'{e}}gas, Pfister, and Wattenberg]{DBLP:conf/nips/0002PVPW23}
Kenneth Li, Oam Patel, Fernanda~B. Vi{\'{e}}gas, Hanspeter Pfister, and Martin Wattenberg.
\newblock Inference-time intervention: Eliciting truthful answers from a language model.
\newblock In Alice Oh, Tristan Naumann, Amir Globerson, Kate Saenko, Moritz Hardt, and Sergey Levine (eds.), \emph{Advances in Neural Information Processing Systems 36: Annual Conference on Neural Information Processing Systems 2023, NeurIPS 2023, New Orleans, LA, USA, December 10 - 16, 2023}, 2023.
\newblock URL \url{http://papers.nips.cc/paper\_files/paper/2023/hash/81b8390039b7302c909cb769f8b6cd93-Abstract-Conference.html}.

\bibitem[Li et~al.(2024)Li, Chen, Chen, Zhang, Su, Xing, and Zhang]{li2024confidencemattersrevisitingintrinsic}
Loka Li, Zhenhao Chen, Guangyi Chen, Yixuan Zhang, Yusheng Su, Eric Xing, and Kun Zhang.
\newblock Confidence matters: Revisiting intrinsic self-correction capabilities of large language models.
\newblock 2024.
\newblock URL \url{https://arxiv.org/abs/2402.12563}.

\bibitem[Ma et~al.(2025)Ma, Chen, Zhou, Wang, and Zhang]{ma2025estimating}
Huan Ma, Jingdong Chen, Joey~Tianyi Zhou, Guangyu Wang, and Changqing Zhang.
\newblock Estimating llm uncertainty with evidence.
\newblock \emph{arXiv preprint arXiv:2502.00290}, 2025.

\bibitem[Mallen et~al.(2023)Mallen, Asai, Zhong, Das, Khashabi, and Hajishirzi]{mallen2022trustworthy}
Alex Mallen, Akari Asai, Victor Zhong, Rajarshi Das, Daniel Khashabi, and Hannaneh Hajishirzi.
\newblock When not to trust language models: Investigating effectiveness of parametric and non-parametric memories.
\newblock In Anna Rogers, Jordan Boyd-Graber, and Naoaki Okazaki (eds.), \emph{Proceedings of the 61st Annual Meeting of the Association for Computational Linguistics (Volume 1: Long Papers)}, pp.\  9802--9822, Toronto, Canada, July 2023. Association for Computational Linguistics.
\newblock \doi{10.18653/v1/2023.acl-long.546}.
\newblock URL \url{https://aclanthology.org/2023.acl-long.546/}.

\bibitem[Manakul et~al.(2023)Manakul, Liusie, and Gales]{manakul2023selfcheckgpt}
Potsawee Manakul, Adian Liusie, and Mark J.~F. Gales.
\newblock Selfcheckgpt: Zero-resource black-box hallucination detection for generative large language models.
\newblock In Houda Bouamor, Juan Pino, and Kalika Bali (eds.), \emph{Proceedings of the 2023 Conference on Empirical Methods in Natural Language Processing, {EMNLP} 2023, Singapore, December 6-10, 2023}, pp.\  9004--9017. Association for Computational Linguistics, 2023.
\newblock \doi{10.18653/V1/2023.EMNLP-MAIN.557}.
\newblock URL \url{https://doi.org/10.18653/v1/2023.emnlp-main.557}.

\bibitem[Obeso et~al.(2025)Obeso, Arditi, Ferrando, Freeman, Holmes, and Nanda]{obeso2025real}
Oscar Obeso, Andy Arditi, Javier Ferrando, Joshua Freeman, Cameron Holmes, and Neel Nanda.
\newblock Real-time detection of hallucinated entities in long-form generation.
\newblock \emph{arXiv preprint arXiv:2509.03531}, 2025.

\bibitem[Orgad et~al.(2024)Orgad, Toker, Gekhman, Reichart, Szpektor, Kotek, and Belinkov]{orgad2024llms}
Hadas Orgad, Michael Toker, Zorik Gekhman, Roi Reichart, Idan Szpektor, Hadas Kotek, and Yonatan Belinkov.
\newblock Llms know more than they show: On the intrinsic representation of llm hallucinations.
\newblock International Conference on Learning Representations (ICLR), 2024.

\bibitem[Qin et~al.(2025)Qin, Li, Nian, Yu, Zhao, and Ma]{qin2025dontlethallucinatepremise}
Yuehan Qin, Shawn Li, Yi~Nian, Xinyan~Velocity Yu, Yue Zhao, and Xuezhe Ma.
\newblock Don't let it hallucinate: Premise verification via retrieval-augmented logical reasoning, 2025.
\newblock URL \url{https://arxiv.org/abs/2504.06438}.

\bibitem[Ravi et~al.(2024)Ravi, Mielczarek, Kannappan, Kiela, and Qian]{lin2024lynx}
Selvan~Sunitha Ravi, Bartosz Mielczarek, Anand Kannappan, Douwe Kiela, and Rebecca Qian.
\newblock Lynx: An open source hallucination evaluation model.
\newblock \emph{arXiv preprint arXiv:2407.08488}, 2024.

\bibitem[Reinhard et~al.(2025)Reinhard, Li, Fina, and Leimeister]{reinhard2025fact}
Philipp Reinhard, Mahei~Manhai Li, Matteo Fina, and Jan~Marco Leimeister.
\newblock Fact or fiction? exploring explanations to identify factual confabulations in rag-based llm systems.
\newblock In \emph{Proceedings of the Extended Abstracts of the CHI Conference on Human Factors in Computing Systems}, CHI EA '25, New York, NY, USA, 2025. Association for Computing Machinery.
\newblock ISBN 9798400713958.
\newblock \doi{10.1145/3706599.3720249}.
\newblock URL \url{https://doi.org/10.1145/3706599.3720249}.

\bibitem[Sensoy et~al.(2018)Sensoy, Kaplan, and Kandemir]{sensoy2018evidentialdeeplearningquantify}
Murat Sensoy, Lance Kaplan, and Melih Kandemir.
\newblock Evidential deep learning to quantify classification uncertainty.
\newblock In \emph{Proceedings of the 32nd International Conference on Neural Information Processing Systems}, NIPS'18, pp.\  3183–3193, Red Hook, NY, USA, 2018. Curran Associates Inc.

\bibitem[Simhi et~al.(2024)Simhi, Herzig, Szpektor, and Belinkov]{simhi2024constructing}
Adi Simhi, Jonathan Herzig, Idan Szpektor, and Yonatan Belinkov.
\newblock Constructing benchmarks and interventions for combating hallucinations in llms.
\newblock \emph{arXiv preprint arXiv:2404.09971}, 2024.

\bibitem[Sriramanan et~al.(2024)Sriramanan, Bharti, Sadasivan, Saha, Kattakinda, and Feizi]{sriramanan2024llm}
Gaurang Sriramanan, Siddhant Bharti, Vinu~Sankar Sadasivan, Shoumik Saha, Priyatham Kattakinda, and Soheil Feizi.
\newblock Llm-check: Investigating detection of hallucinations in large language models.
\newblock \emph{Advances in Neural Information Processing Systems}, 37:\penalty0 34188--34216, 2024.

\bibitem[Sui et~al.(2024)Sui, Duede, Wu, and So]{sui2024confabulationsurprisingvaluelarge}
Peiqi Sui, Eamon Duede, Sophie Wu, and Richard So.
\newblock Confabulation: The surprising value of large language model hallucinations.
\newblock In Lun-Wei Ku, Andre Martins, and Vivek Srikumar (eds.), \emph{Proceedings of the 62nd Annual Meeting of the Association for Computational Linguistics (Volume 1: Long Papers)}, pp.\  14274--14284, Bangkok, Thailand, August 2024. Association for Computational Linguistics.
\newblock \doi{10.18653/v1/2024.acl-long.770}.
\newblock URL \url{https://aclanthology.org/2024.acl-long.770/}.

\bibitem[Team et~al.(2025)Team, Abbas, Ahmad, Alam, Altinisik, Asgari, Boshmaf, Boughorbel, Chawla, Chowdhury, et~al.]{fanarteam2025fanararabiccentricmultimodalgenerative}
Fanar Team, Ummar Abbas, Mohammad~Shahmeer Ahmad, Firoj Alam, Enes Altinisik, Ehsannedin Asgari, Yazan Boshmaf, Sabri Boughorbel, Sanjay Chawla, Shammur Chowdhury, et~al.
\newblock Fanar: An arabic-centric multimodal generative ai platform.
\newblock \emph{arXiv preprint arXiv:2501.13944}, 2025.

\bibitem[Wang et~al.(2023)Wang, Wei, Schuurmans, Bosma, Chi, and Zhou]{wang2023selfconsistency}
Xuezhi Wang, Jason Wei, Dale Schuurmans, Maarten Bosma, Ed~Chi, and Denny Zhou.
\newblock Self-consistency improves chain of thought reasoning in language models.
\newblock In \emph{International Conference on Learning Representations (ICLR)}, 2023.

\bibitem[Wei et~al.(2022)Wei, Wang, Schuurmans, Bosma, Ichter, Xia, Chi, Le, and Zhou]{wei2022chain}
Jason Wei, Xuezhi Wang, Dale Schuurmans, Maarten Bosma, Brian Ichter, Fei Xia, Ed~H. Chi, Quoc~V. Le, and Denny Zhou.
\newblock Chain-of-thought prompting elicits reasoning in large language models.
\newblock In \emph{Proceedings of the 36th International Conference on Neural Information Processing Systems}, NIPS '22, Red Hook, NY, USA, 2022. Curran Associates Inc.
\newblock ISBN 9781713871088.

\bibitem[Wei et~al.(2024)Wei, Yang, Song, Lu, Hu, Huang, Tran, Peng, Liu, Huang, Du, and Le]{wei2024longformfactualitylargelanguage}
Jerry Wei, Chengrun Yang, Xinying Song, Yifeng Lu, Nathan Hu, Jie Huang, Dustin Tran, Daiyi Peng, Ruibo Liu, Da~Huang, Cosmo Du, and Quoc~V. Le.
\newblock Long-form factuality in large language models.
\newblock In \emph{Proceedings of the 38th International Conference on Neural Information Processing Systems}, NIPS '24, Red Hook, NY, USA, 2024. Curran Associates Inc.
\newblock ISBN 9798331314385.

\bibitem[Wu et~al.(2022)Wu, Zeng, He, Mou, Wang, and Xu]{wu2022distributioncalibrationoutofdomaindetection}
Yanan Wu, Zhiyuan Zeng, Keqing He, Yutao Mou, Pei Wang, and Weiran Xu.
\newblock Distribution calibration for out-of-domain detection with {B}ayesian approximation.
\newblock In Nicoletta Calzolari, Chu-Ren Huang, Hansaem Kim, James Pustejovsky, Leo Wanner, Key-Sun Choi, Pum-Mo Ryu, Hsin-Hsi Chen, Lucia Donatelli, Heng Ji, Sadao Kurohashi, Patrizia Paggio, Nianwen Xue, Seokhwan Kim, Younggyun Hahm, Zhong He, Tony~Kyungil Lee, Enrico Santus, Francis Bond, and Seung-Hoon Na (eds.), \emph{Proceedings of the 29th International Conference on Computational Linguistics}, pp.\  608--615, Gyeongju, Republic of Korea, October 2022. International Committee on Computational Linguistics.
\newblock URL \url{https://aclanthology.org/2022.coling-1.50/}.

\bibitem[Yadkori et~al.(2024{\natexlab{a}})Yadkori, Kuzborskij, Gy\"{o}rgy, and Szepesv\'{a}ri]{Yadkori_Kuzborskij_György_Szepesvári_2024}
Yasin~Abbasi Yadkori, Ilja Kuzborskij, Andr\'{a}s Gy\"{o}rgy, and Csaba Szepesv\'{a}ri.
\newblock To believe or not to believe your llm: Iterative prompting for estimating epistemic uncertainty.
\newblock In \emph{Proceedings of the 38th International Conference on Neural Information Processing Systems}, NIPS '24, Red Hook, NY, USA, 2024{\natexlab{a}}. Curran Associates Inc.
\newblock ISBN 9798331314385.

\bibitem[Yadkori et~al.(2024{\natexlab{b}})Yadkori, Kuzborskij, Stutz, György, Fisch, Doucet, Beloshapka, Weng, Yang, Szepesvári, Cemgil, and Tomasev]{Yadkori_Kuzborskij_Stutz_György_Fisch_Doucet_Beloshapka_Weng_Yang_Szepesvári_et2024}
Yasin~Abbasi Yadkori, Ilja Kuzborskij, David Stutz, András György, Adam Fisch, Arnaud Doucet, Iuliya Beloshapka, Wei-Hung Weng, Yao-Yuan Yang, Csaba Szepesvári, Ali~Taylan Cemgil, and Nenad Tomasev.
\newblock Mitigating llm hallucinations via conformal abstention.
\newblock \penalty0 (arXiv:2405.01563), April 2024{\natexlab{b}}.
\newblock \doi{10.48550/arXiv.2405.01563}.
\newblock URL \url{http://arxiv.org/abs/2405.01563}.
\newblock arXiv:2405.01563 [cs].

\bibitem[Yang et~al.(2018)Yang, Qi, Zhang, Bengio, Cohen, Salakhutdinov, and Manning]{Yang0ZBCSM18}
Zhilin Yang, Peng Qi, Saizheng Zhang, Yoshua Bengio, William~W. Cohen, Ruslan Salakhutdinov, and Christopher~D. Manning.
\newblock Hotpotqa: {A} dataset for diverse, explainable multi-hop question answering.
\newblock In \emph{Proceedings of the 2018 Conference on Empirical Methods in Natural Language Processing, Brussels, Belgium, October 31 - November 4, 2018}, pp.\  2369--2380. Association for Computational Linguistics, 2018.
\newblock \doi{10.18653/V1/D18-1259}.
\newblock URL \url{https://doi.org/10.18653/v1/d18-1259}.

\bibitem[Zhang et~al.(2024)Zhang, Juan, Rashtchian, Ferng, Jiang, and Chen]{zhang2025sledselflogitsevolution}
Jianyi Zhang, Da-Cheng Juan, Cyrus Rashtchian, Chun-Sung Ferng, Heinrich Jiang, and Yiran Chen.
\newblock Sled: self logits evolution decoding for improving factuality in large language models.
\newblock In \emph{Proceedings of the 38th International Conference on Neural Information Processing Systems}, NIPS '24, Red Hook, NY, USA, 2024. Curran Associates Inc.
\newblock ISBN 9798331314385.

\bibitem[Şensoy et~al.(2025)Şensoy, Kaplan, Julier, Saleki, and Cerutti]{Sensoy_Kaplan_Julier_Saleki_Cerutti_2025}
Murat Şensoy, Lance~M. Kaplan, Simon Julier, Maryam Saleki, and Federico Cerutti.
\newblock Risk-aware classification via uncertainty quantification.
\newblock \emph{Expert Systems with Applications}, 265:\penalty0 125906, March 2025.
\newblock ISSN 09574174.
\newblock \doi{10.1016/j.eswa.2024.125906}.

\end{thebibliography}

\appendix

\section{Implementation Details}
\label{ap:implementation-details}

\subsection{Prompts for Different Experiments}
\paragraph{Response Generation}
Since our datasets consist of direct QA pairs without elaboration, we prompt the LLM to answer questions in the same concise manner. This ensures alignment with the ground truth format and allows for fair comparison across model outputs.~\begin{tcolorbox}[]
    \texttt{Answer the question directly, without additional explanation, and be as concise as possible.} \\
\end{tcolorbox}

\paragraph{Incorrect Context Generation}
To support the \texttt{WIC} experimental condition, we use GPT-4.1 mini to generate misleading but plausible context for each question. This allows us to simulate scenarios in which the LLM is exposed to confounding information, enabling evaluation of its susceptibility to plausible but incorrect cues.~\begin{tcolorbox}[]
    System Prompt:\\
    \texttt{You are an incorrect context generator. Given a question Q, generate a short made up context information that misleads the question from
    giving a correct answer. Make sure your context information does not lead to the correct answer A but rather lead to an incorrect but seemingly correct response.}\\
    User Prompt:\\
    \texttt{Q: [Question]} \\
    \texttt{A: [Answer]}\\
\end{tcolorbox}
We apply this prompt to the subset of question–response pairs that were consistently answered correctly under the \texttt{WOC} setting. The goal is to inject misleading context into otherwise confidently answered questions in order to analyze how model uncertainty behaves under deceptive conditions.

\paragraph{RAG Context Injection}
We simulate a real-world Retrieval-Augmented Generation (RAG) system by adopting a prompt adapted from Azure’s official RAG documentation\footnote{https://learn.microsoft.com/en-us/azure/search/tutorial-rag-build-solution-pipeline}. This prompt constrains the LLM to generate responses strictly based on the provided sources, enabling us to assess whether the model can produce accurate and well-grounded answers when external context is explicitly injected.~\begin{tcolorbox}[]

    \texttt{You are an AI assistant that helps users learn from the information found in the source material.} \\
    \texttt{Answer the query concisely using only the sources provided below.}\\
    \texttt{If the answer is longer than 3 sentences, provide a summary.}\\
    \texttt{Answer ONLY with the facts listed in the list of sources below. Cite your source when you answer the question.}\\
    \texttt{If there isn't enough information below, say you don't know.}\\
    \texttt{Do not generate answers that don't use the sources below.}\\
    \texttt{Answer the question directly, without additional explanation, and be as concise as possible. Use maximum 15 words in your response.}\\
    \texttt{Query: [Query]}\\
    \texttt{Sources:[Sources]}\\
\end{tcolorbox}

\paragraph{LLM as a Judge}
Because ground truth correctness labels are absent in our datasets and manual annotation is resource-intensive, we use an LLM-as-a-judge approach. Prior research shows this method closely approximates human judgment, making it suitable for generating labels used in AUROC scoring.~\begin{tcolorbox}[]
    \texttt{Given a question and a ground truth answer, judge the correctness of the candidate response.}\\
    \texttt{**Important Definitions**:}\\
    \texttt{- A response is considered **correct** if it matches the **key information** of the ground truth answer.}\\
    \texttt{- A response is **incorrect** if it is factually wrong, off-topic, or misleading.}\\
    \texttt{Return 1 if correct, return 0 if incorrect. Do not return anything else.}
\end{tcolorbox}

\subsection{Baseline Implementation}

\paragraph{P(True):} We follow the implementation following template from \citet{kadavath2022language}. We prompt each LLM to judge their responses following the prompt template:
\begin{tcolorbox}[]
    \texttt{Question: [Question]} \\
    \texttt{Proposed Answer: [LLM long answer]}\\
    \texttt{Is the proposed answer:} \\
    \texttt{(1) True} \\
    \texttt{(0) False} \\
    \texttt{The proposed answer is:}
\end{tcolorbox}

\paragraph{LogTokU:}
LogTokU \cite{ma2025estimating} enhances response-level reliability estimation by addressing the tendency of probability-based methods to overestimate uncertainty for uninformative tokens, such as punctuation. Unlike entropy-based approaches that require heuristic reweighting to downplay these tokens, LogTokU leverages token-level uncertainty quadrants to naturally separate informative from ambiguous tokens. Following this formulation, we compute response reliability as the average reliability over the K least reliable tokens, 
\begin{equation}
\mathcal{R}_{\text{response}} = \frac{1}{K} \sum_{t \in \mathcal{T}_K} \mathcal{R}(a_t),
\end{equation}
where each token’s reliability is defined as:
\begin{equation}
\text{Reliability} = - \text{AU} \cdot \text{EU}
\label{eq:reliability}
\end{equation}

Here, AU and EU represent aleatoric and epistemic uncertainty, respectively, as defined in Equations~\ref{eq:au} and~\ref{eq:eu}. In our experiments, we set K=10.

Additionally, we explored a variant, LogTokU (imp), where instead of selecting tokens solely based on low reliability, we aggregate AU and EU scores over semantically important tokens in the response. This approach tests whether focusing on key content-bearing tokens provides a more faithful reliability estimate.

\paragraph{Probing: } We describe the probing-based reliability estimation methods in greater detail. Following prior work~\citep{DBLP:conf/nips/0002PVPW23}, we train lightweight logistic regression classifiers on token-level hidden states extracted from the language model to predict binary response-level reliability labels.

Given a generated response $\mathbf{y} = (y_1, \ldots, y_T)$ and the corresponding hidden state vectors $\mathbf{h}^{(l)}_1, \ldots, \mathbf{h}^{(l)}_T$ at layer $l$, we explore multiple token selection strategies to construct input features for the classifier:

\begin{itemize}
    \item \textbf{Probe(EOS).} We use the hidden state of the final token $\mathbf{h}^{(l)}_{T}$, which is often used in autoregressive decoding as a summary representation.

    \item \textbf{Probe(Exact).} Inspired by the span-based alignment procedure used in~\citep{orgad2024llms}. We select the hidden states of tokens that align with the exact answer span in the output. If the ground truth answer is multi-token, we average the hidden states of all matching tokens. More concretely, we prompt the ChatGPT-4o to extract the excat anwer tokens. We use the following prompt
    
        \begin{tcolorbox}[] 
        \textbf{System Prompt (Factual Evaluation Task)} \\[4pt]
        You are an expert factual evaluator. Your task is to evaluate whether a given \textbf{Response} to a \textbf{Question} is factually correct based on the provided ground truth \textbf{Answer}. You should do the following: 
        \begin{itemize}

            \item \textbf{Correctness}: Determine whether the LLM's answer is factually correct based on the provided ground truth. An answer is correct if it contains or conveys the correct answer unambiguously. Output a binary label:
            \begin{itemize}
                \item \texttt{"label": 1} \quad if the answer is correct
                \item \texttt{"label": 0} \quad if the answer is incorrect
            \end{itemize}
        
            \item \textbf{Response Extraction}: Regardless of correctness, extract the \textbf{minimal, meaningful tokens} from the \textbf{Response} that attempt to directly answer the question. This is the part of the response that the model presents as the main answer (even if it is wrong or uncertain). Extract no more than 3 tokens. Use the \textbf{Question} and \textbf{Answer} to infer which part of the \textbf{Response} is the most relevant.
        \end{itemize}
        
        You must return your output as a dictionary in the format:
\texttt{\{"label": 0 or 1, "exact\_answer": "substring from Response"\}}

        \end{tcolorbox}

    \item \textbf{Probe(EU).} We compute epistemic uncertainty (EU) scores for each token using Equation \ref{eq:eu}. We then select the hidden state of the token with either the highest or lowest EU, under the hypothesis that these tokens are most indicative of reliability.

    \item \textbf{Probe(AVG).} Instead of selecting a single token, we average the hidden states across a subset of tokens to construct a fixed-length feature vector. The candidate subsets include: (1) the top-$k$ most uncertain or certain tokens, as measured by epistemic uncertainty (EU), and (2) fixed heuristic positions, such as the first and last tokens in the generated response. The aggregated hidden state vector is then used as input to the classifier. We evaluate all subset strategies and report the performance of the best-performing one as the final result for \textbf{Probe(AVG)}.

\end{itemize}

All classifiers are trained on a small held-out portion of labeled data using 70/30 train-test splits and evaluated using accuracy and AUROC.

\section{Additional Experimental Results}
\label{ap:additional-experimental-results}


\begin{figure*}[!t]
    \centering
    \begin{subfigure}[t]{0.47\textwidth}
        \centering
        \includegraphics[width=\linewidth]{figures/rq1_eu_meanlow.pdf}
        \caption{Top-$10$ lowest EU mean}
    \end{subfigure}
    \hfill
    \begin{subfigure}[t]{0.47\textwidth}
        \centering
        \includegraphics[width=\linewidth]{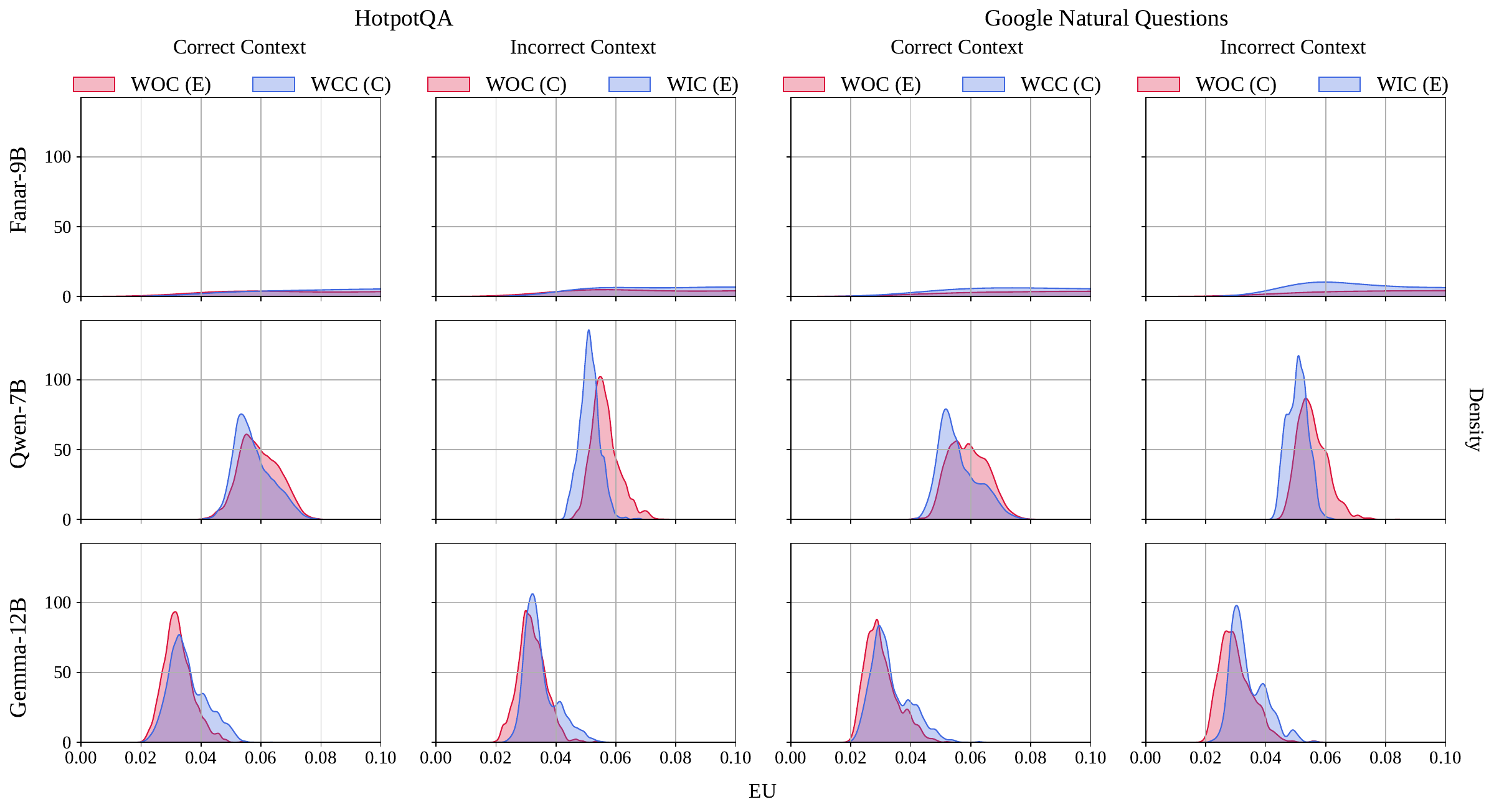}
        \caption{Top-$10$ highest EU mean}
    \end{subfigure}

    \vspace{0.5em}

    \begin{subfigure}[t]{0.47\textwidth}
        \centering
        \includegraphics[width=\linewidth]{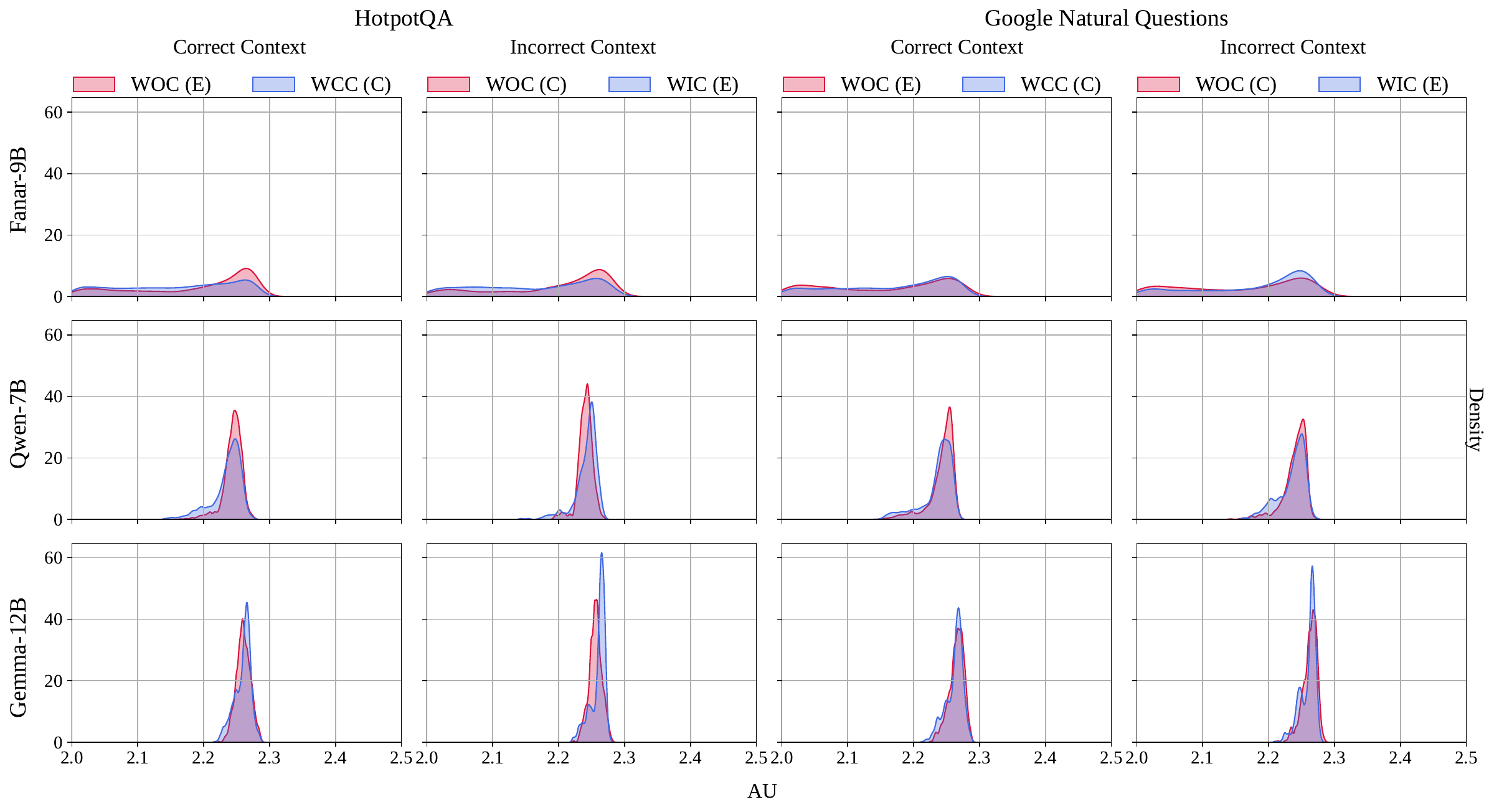}
        \caption{Top-$10$ lowest AU mean}
    \end{subfigure}
    \hfill
    \begin{subfigure}[t]{0.47\textwidth}
        \centering
        \includegraphics[width=\linewidth]{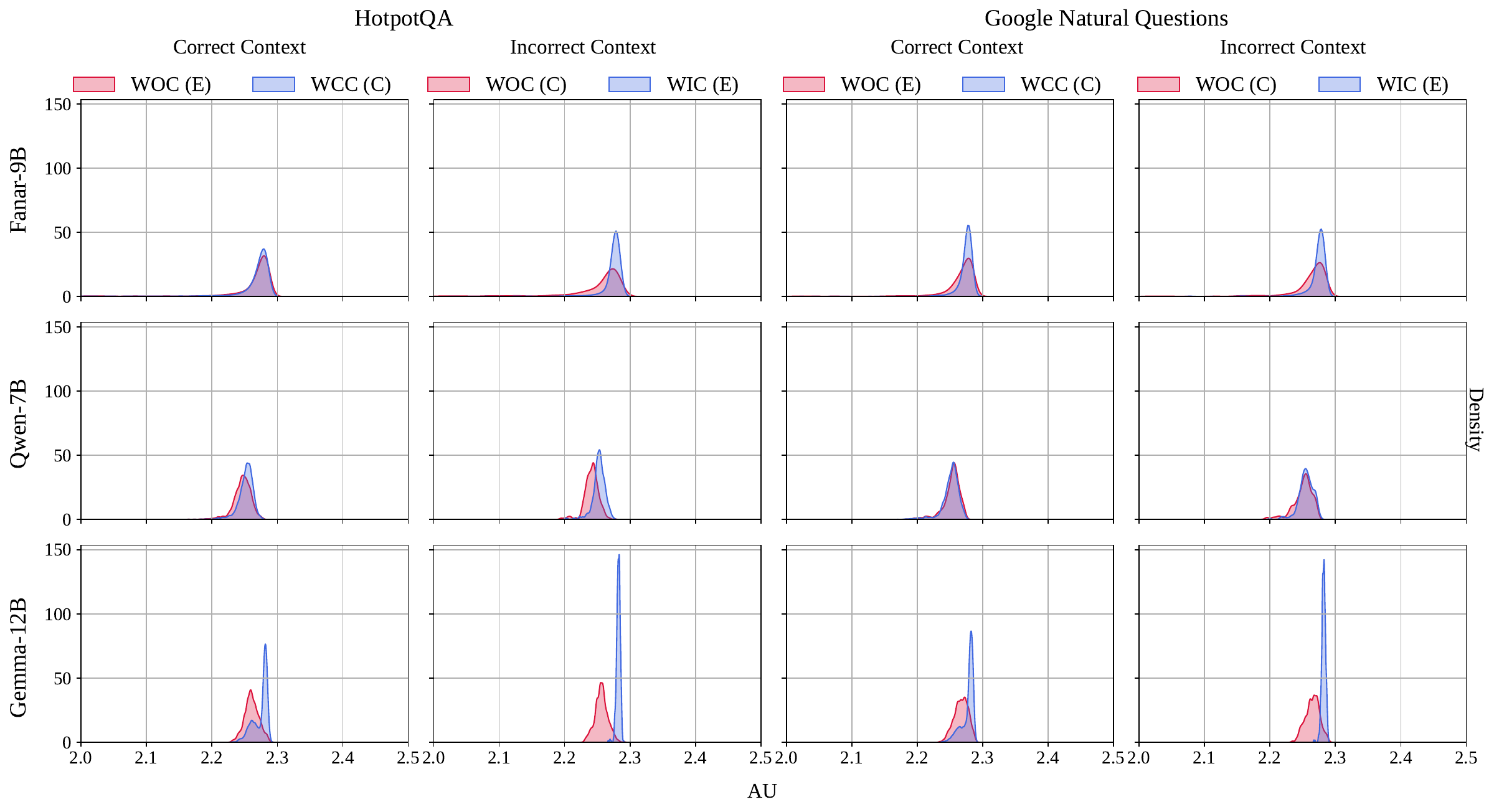}
        \caption{Top-$10$ highest AU mean}
    \end{subfigure}

    \vspace{0.5em}

    \begin{subfigure}[t]{0.47\textwidth}
        \centering
        \includegraphics[width=\linewidth]{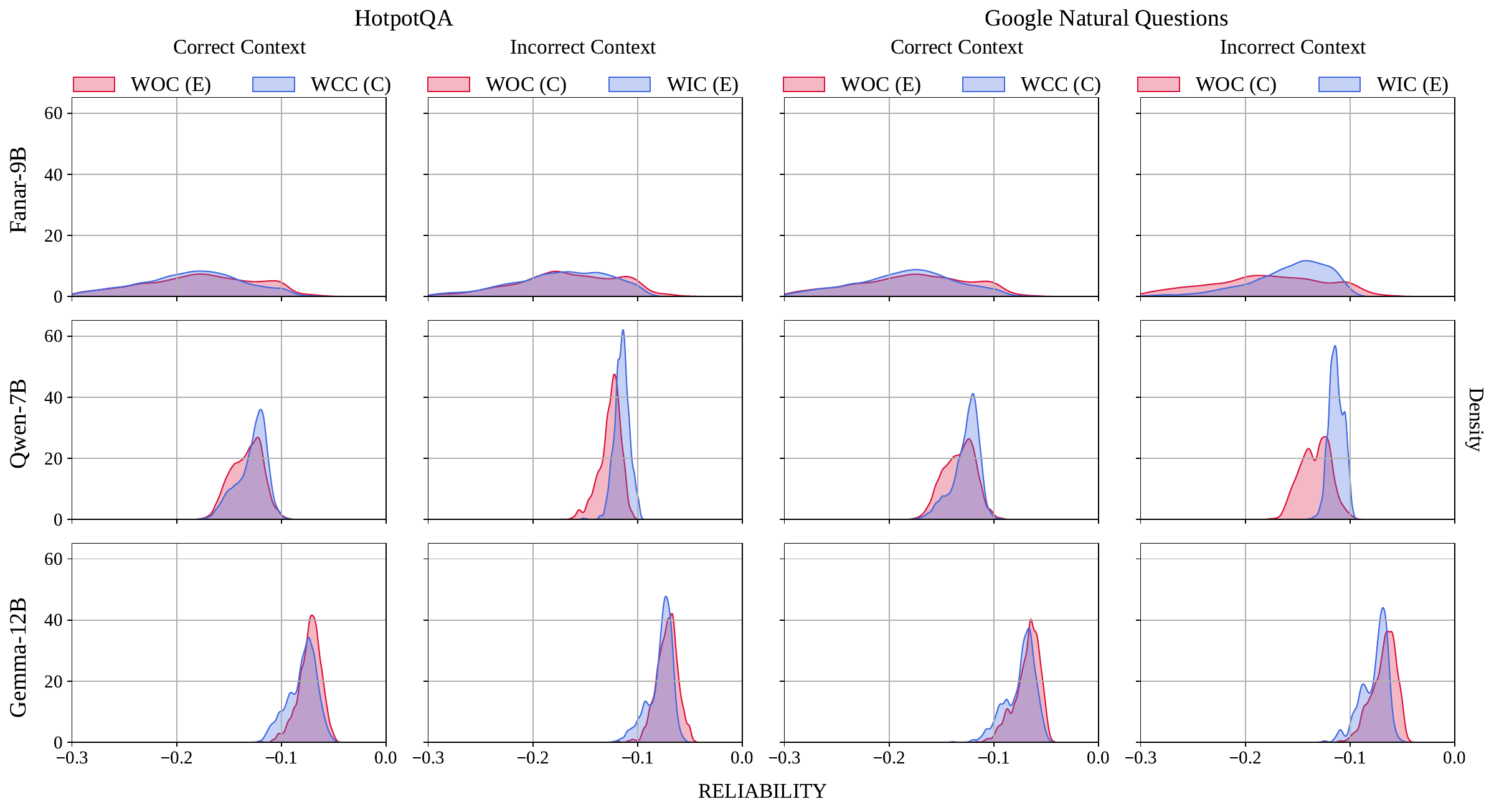}
        \caption{Top-$10$ lowest reliability (most unreliable tokens)}
    \end{subfigure}
    \hfill
    \begin{subfigure}[t]{0.47\textwidth}
        \centering
        \includegraphics[width=\linewidth]{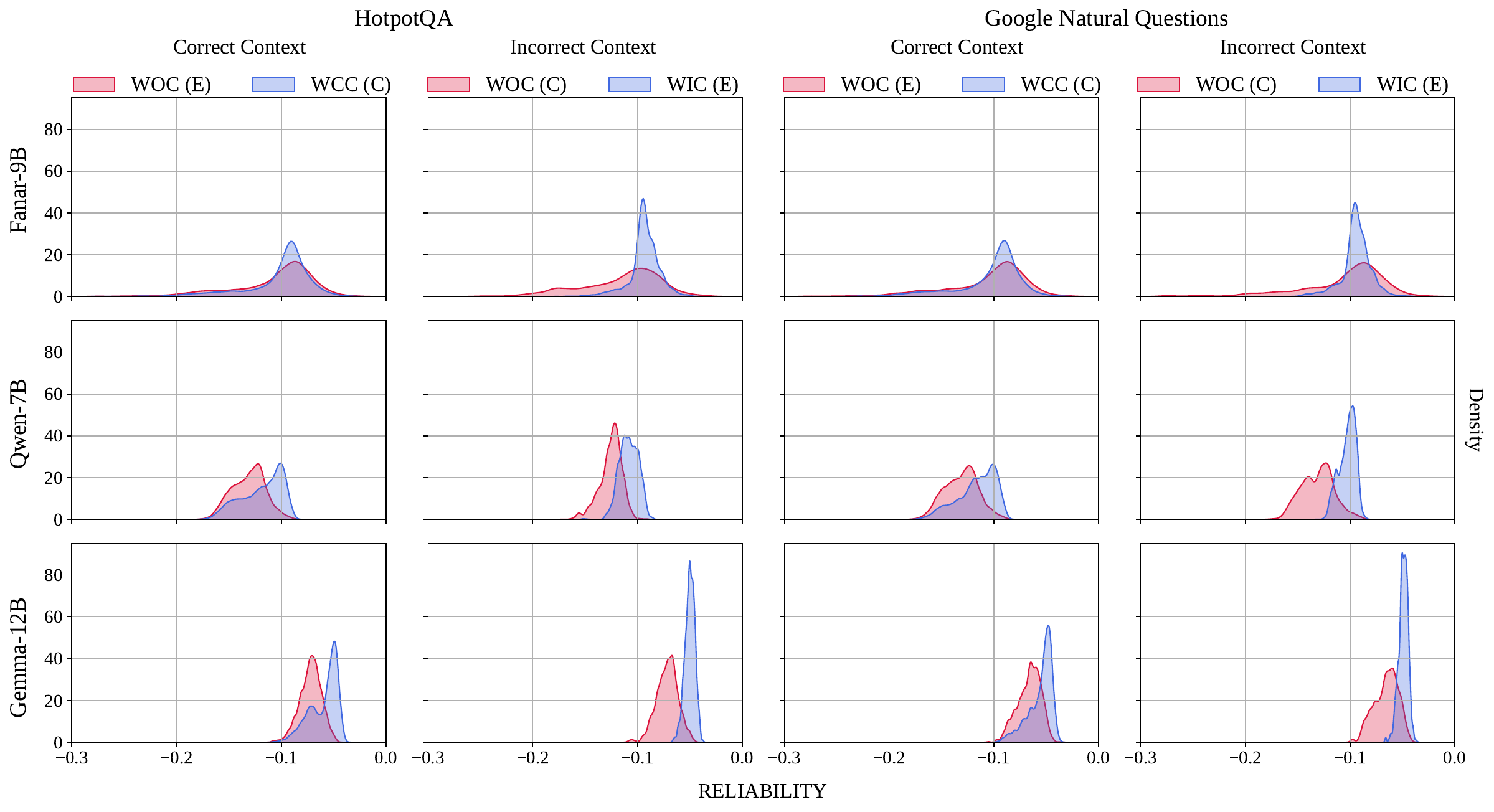}
        \caption{Top-$10$ highest reliability (most reliable tokens)}
    \end{subfigure}

    \caption{We analyze the transitions in error types and shifts in uncertainty distributions across the \hotpot and \NQ datasets for three models (\fanar, \qwen, \gemma). Our evaluation considers three uncertainty measures: epistemic uncertainty, aleatoric uncertainty, and a composite reliability score. Among the examined features, the mean of the top-$K$ lowest epistemic uncertainty (EU) scores—using $K=10$—proves to be the most indicative. This finding supports our hypothesis that incorporating external context not only reduces model uncertainty but also decreases the variance across predictions.}
    \label{fig:six-figures}
\end{figure*}

\begin{figure*}[ht]
\centering
\includegraphics[width=\textwidth]{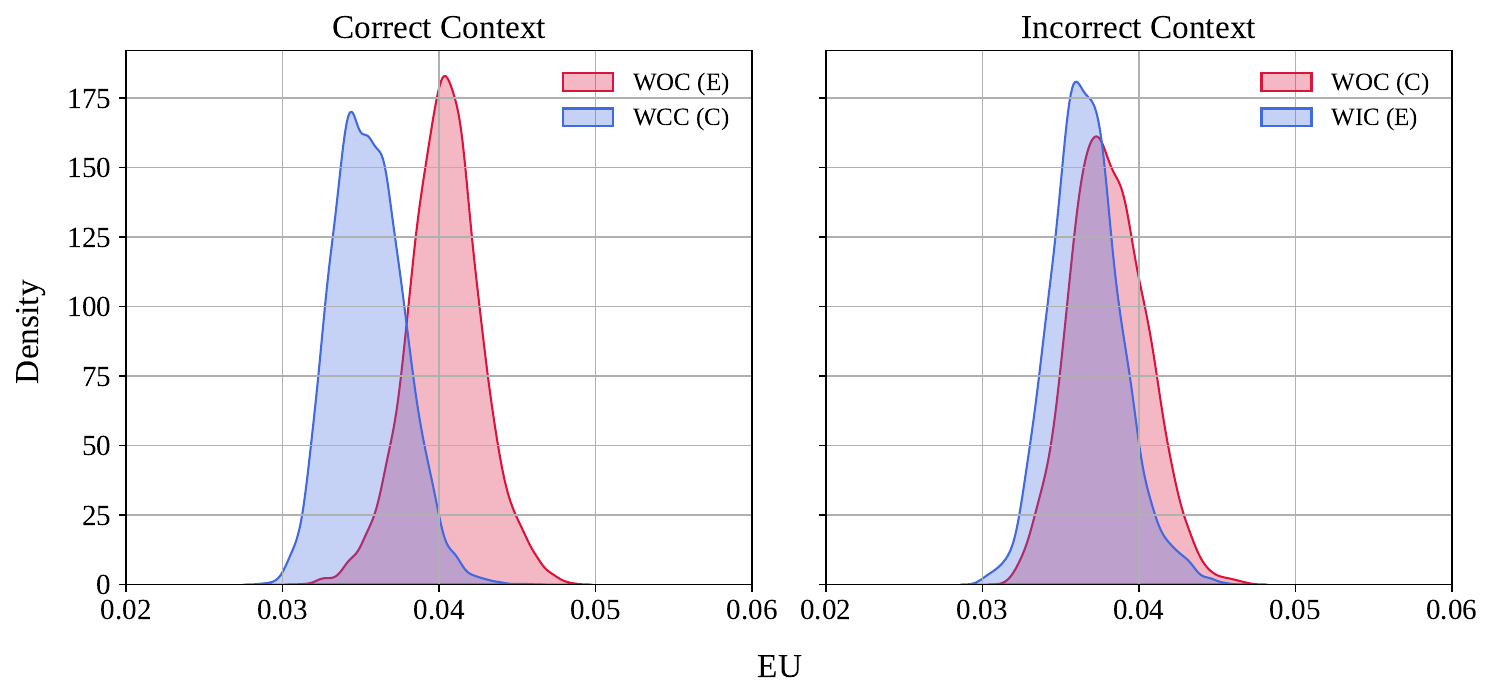}
\caption{Lower EU mean distributions for the \NQ dataset using the \oss model, evaluated under the same experimental setup as Figure~\ref{fig:eu_shift}. The results align with those observed for \fanar, \qwen, and \gemma, showing the expected leftward shift in \texttt{WOC:E}$\rightarrow$\texttt{WCC:C} and sharper distributions in both transitions. For \texttt{WOC:C}$\rightarrow$\texttt{WIC:E}, \oss displays relatively stable EU compared to the other models, suggesting improved calibration when misleading context is introduced.}
\label{fig:oss_nq}
\end{figure*}

\begin{figure*}[ht]
\centering
\includegraphics[width=\textwidth]{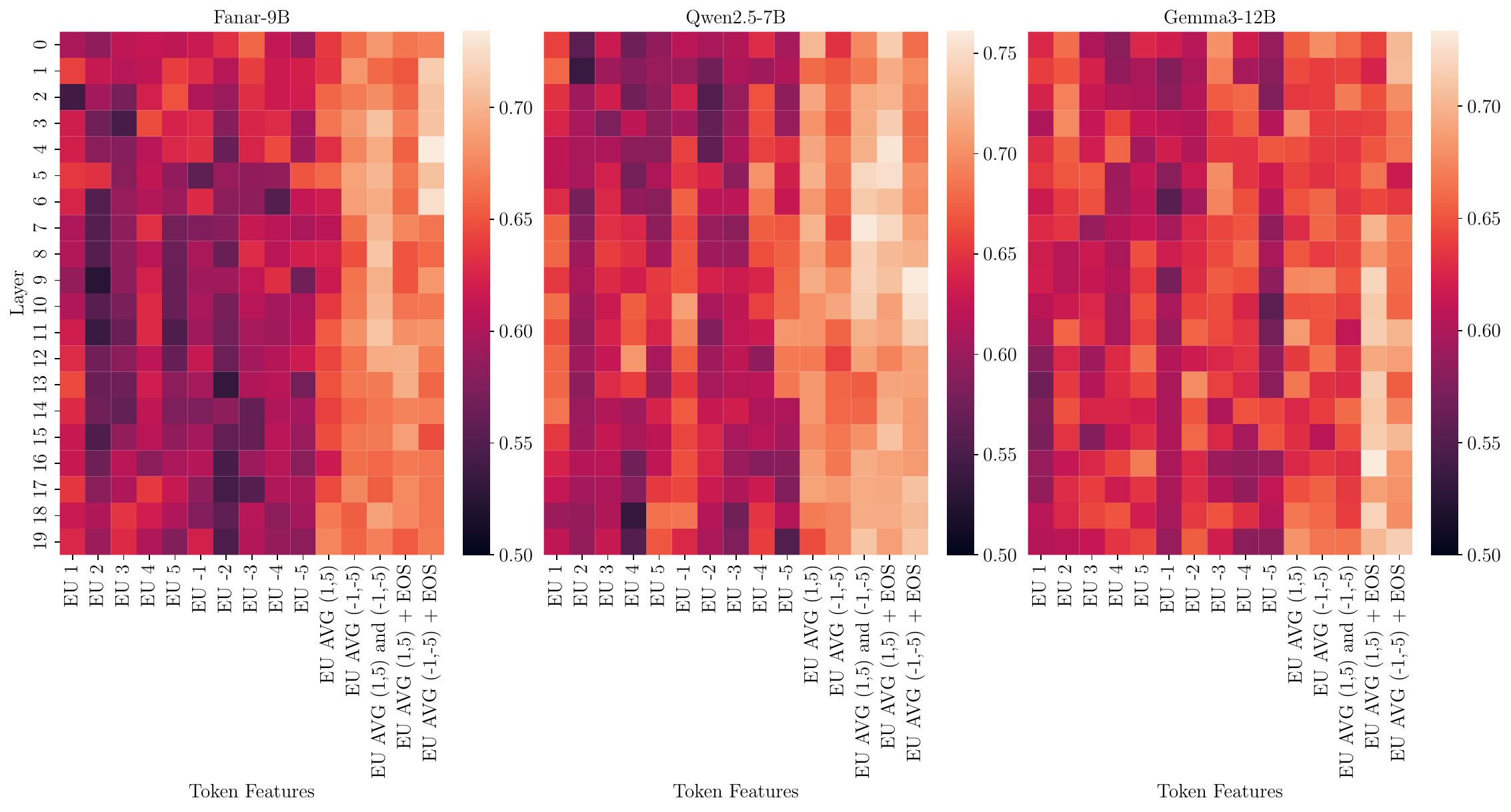}
\caption{AUROC scores of probing classifiers across the last 20 layers using different token-level features, evaluated on the \truthfulqa dataset for \fanar, \qwen, and \gemma. From left to right, columns correspond to probing with single tokens ranked by epistemic uncertainty: the $k$ smallest (EU 1 to EU 5) and the $k$ largest (EU -1 to EU -5). Aggregated features (EU AVG) formed by averaging hidden states across selected tokens yield the highest detection performance across all models.}
\label{fig:token_feature_heatmap}
\end{figure*}

\begin{figure*}[ht]
\centering
\includegraphics[width=\textwidth]{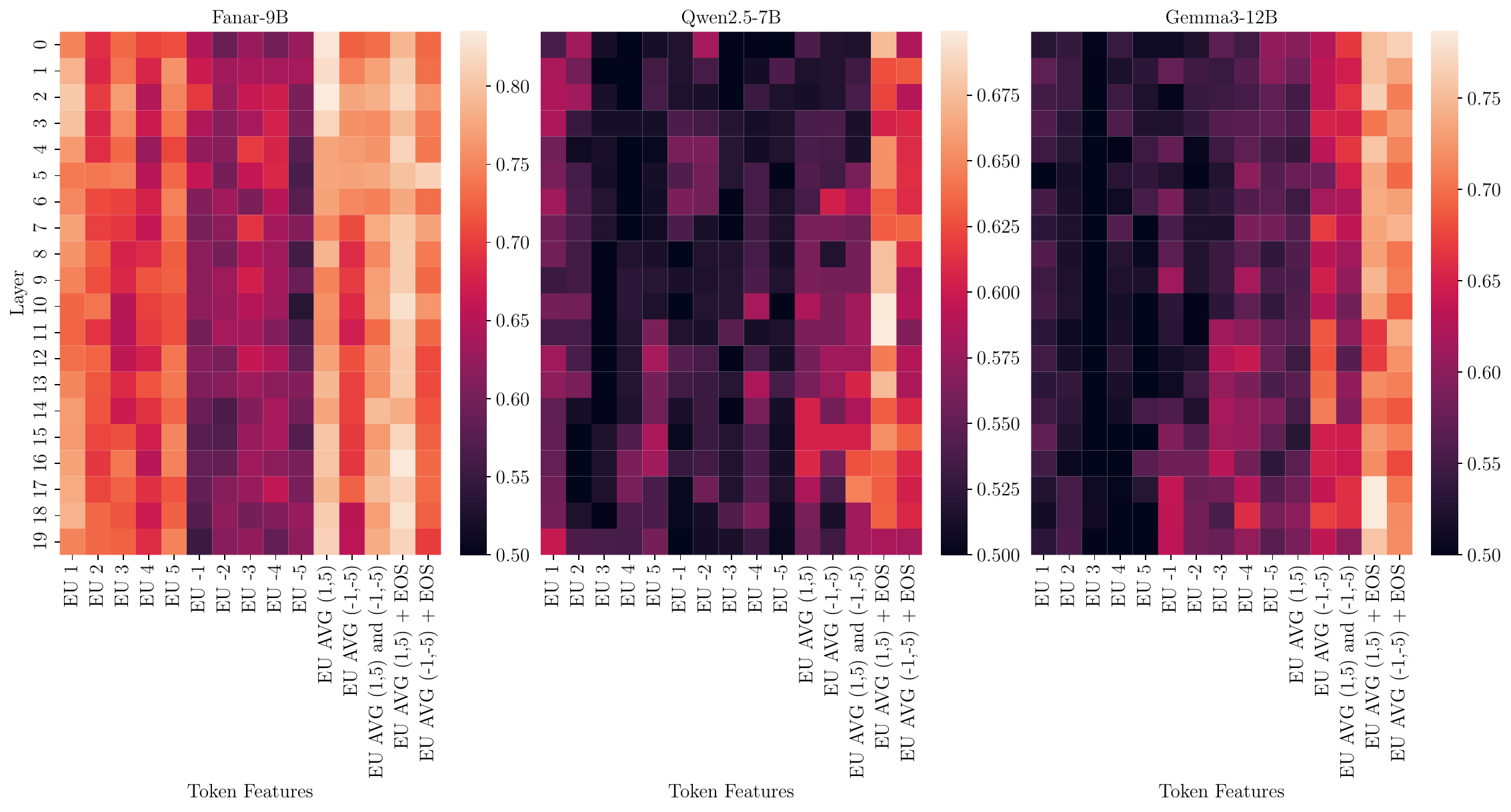}
\caption{AUROC scores of probing classifiers across the last 20 layers using different token-level features, evaluated on the \mathqa dataset for \fanar, \qwen, and \gemma. From left to right, columns correspond to probing with single tokens ranked by epistemic uncertainty: the $k$ smallest (EU 1 to EU 5) and the $k$ largest (EU -1 to EU -5). Aggregated features (EU AVG) formed by averaging hidden states across selected tokens yield the highest detection performance across all models.}
\label{fig:token_feature_heatmap}
\end{figure*}

\end{document}